\documentclass{article}

\usepackage{natbib}
\setcitestyle{numbers, compress}
\usepackage{lipsum}
\usepackage{titletoc}
\usepackage{minitoc}
\usepackage{microtype}
\usepackage{graphicx}
\usepackage{booktabs} 
\usepackage{xcolor}         
\usepackage{wrapfig}
\usepackage{pifont}

\usepackage[pagebackref]{hyperref}  

\hypersetup{colorlinks=true,linkcolor=red!70!black,linktocpage=false,citebordercolor=blue!70!black,citecolor=blue!70!black,anchorcolor=blue!70!black}

\usepackage[accepted]{icml2025}

\usepackage{amsmath,amsfonts,bm}

\def\1{\bm{1}}

\def\rF{{\textnormal{F}}}

\DeclareMathAlphabet{\mathsfit}{\encodingdefault}{\sfdefault}{m}{sl}
\SetMathAlphabet{\mathsfit}{bold}{\encodingdefault}{\sfdefault}{bx}{n}

\newcommand{\sm}{{\rm softmax}}

\DeclareMathOperator{\Tr}{Tr}

\newcommand{\SA}{{\textnormal{\textsf{SA}}}}
\newcommand{\TF}{{\textnormal{\textsf{TF}}}}
\newcommand{\FFN}{\textnormal{\textsf{FFN}}}
\newcommand{\LN}{\textnormal{\textsf{Norm}}}
\newcommand{\FFFN}{{\textnormal{\textsf{F}}}}
\newcommand{\MSA}{{\textnormal{\textsf{M}}}}
\newcommand{\ASA}{{\textnormal{\textsf{A}}}}
\newcommand{\SSA}{{\textnormal{\textsf{S}}}}
\newcommand{\Embed}{\textnormal{\textsf{Emb}}}
\newcommand{\QK}{\textnormal{\textsf{QK}}}
\newcommand{\VO}{\textnormal{\textsf{VO}}}
\newcommand{\Norm}{\textnormal{\textsf{Norm}}}

\newcommand{\trinorm}[1]{{\left\vert\kern-0.25ex\left\vert\kern-0.25ex\left\vert #1 
   \right\vert\kern-0.25ex\right\vert\kern-0.25ex\right\vert}}

\newcommand{\ba}{\boldsymbol{a}}

\newcommand{\bh}{\boldsymbol{h}}

\newcommand{\bw}{\boldsymbol{w}}
\newcommand{\bx}{\boldsymbol{x}}
\newcommand{\by}{\boldsymbol{y}}
\newcommand{\bz}{\boldsymbol{z}}
\newcommand{\bA}{\boldsymbol{A}}
\newcommand{\bB}{\boldsymbol{B}}

\newcommand{\bI}{\boldsymbol{I}}

\newcommand{\bK}{\boldsymbol{K}}

\newcommand{\bM}{\boldsymbol{M}}

\newcommand{\bW}{\boldsymbol{W}}
\newcommand{\bX}{\boldsymbol{X}}
\newcommand{\bY}{\boldsymbol{Y}}
\newcommand{\bZ}{\boldsymbol{Z}}

\newcommand{\bbeta}{\bm{\beta}}

\newcommand{\bgamma}{\bm{\gamma}}

\newcommand{\btheta}{\boldsymbol{\theta}}

\newcommand{\cL}{\mathcal{L}}

\newcommand{\cO}{\mathcal{O}}

\newcommand{\cQ}{\mathcal{Q}}

\newcommand{\cS}{\mathcal{S}}

\newcommand{\bbE}{\mathbb{E}}

\newcommand{\bbR}{\mathbb{R}}

\newcommand{\bbV}{\mathbb{V}}

\newcommand{\bbZ}{\mathbb{Z}}

\newcommand{\bone}{\mathbf{1}}
\newcommand{\pll}{\kern 0.56em/\kern -0.8em /\kern 0.56em}

\newcommand{\norm}[1]{\ensuremath{\left\| #1 \right\|}}
\newcommand{\bracket}[1]{\ensuremath{\left( #1 \right)}}

\newcommand{\diag}{{\rm diag}}

\renewcommand{\vec}{{\rm vec}}
\newcommand{\<}{\left\langle}
\renewcommand{\>}{\right\rangle}

\usepackage{graphicx,subfig}
\usepackage{url}
\usepackage{multirow}
\usepackage{amsmath}
\usepackage{amssymb}
\usepackage{mathtools}
\usepackage{amsthm}
\usepackage{enumitem}
\usepackage{stfloats}
\usepackage{makecell}
\usepackage{tcolorbox}
\usepackage{framed}
\colorlet{shadecolor}{orange!15}

\usepackage[capitalize,noabbrev]{cleveref}

\usepackage[textsize=tiny]{todonotes}

\theoremstyle{plain}
\newtheorem{theorem}{Theorem}[section]
\newtheorem{remark}[theorem]{Remark}

\theoremstyle{definition}

\allowdisplaybreaks[4]

\usepackage[textsize=tiny]{todonotes}

\icmltitlerunning{The Sharpness Disparity Principle in Transformers for Accelerating Language Model Pre-Training}

\begin{document}

\twocolumn[


\icmltitle{The Sharpness Disparity Principle in Transformers \\ for Accelerating Language Model Pre-Training}




\icmlsetsymbol{equal}{*}

\begin{icmlauthorlist}
\icmlauthor{Jinbo Wang}{equal,sms}
\icmlauthor{Mingze Wang}{equal,sms}
\icmlauthor{Zhanpeng Zhou}{equal,scs}
\icmlauthor{Junchi Yan}{scs}
\icmlauthor{Weinan E}{sms,cmlr,aisi}
\icmlauthor{Lei Wu}{sms,cmlr,aisi}
\end{icmlauthorlist}

\icmlaffiliation{sms}{School of Mathematical Sciences, Peking University}
\icmlaffiliation{cmlr}{Center for Machine Learning Research, Peking University}
\icmlaffiliation{scs}{Shanghai Jiao Tong University}
\icmlaffiliation{aisi}{AI for Science Institute, Beijing, China}

\icmlcorrespondingauthor{Lei Wu}{leiwu@math.pku.edu.cn}
\icmlcorrespondingauthor{Mingze Wang}{mingzewang@stu.pku.edu.cn}

\icmlkeywords{Machine Learning, ICML}

\vskip 0.3in
]



\printAffiliationsAndNotice{\icmlEqualContribution} 

\begin{abstract}
Transformers consist of diverse building blocks, such as embedding layers, normalization layers, self-attention mechanisms, and point-wise feedforward networks. Thus, understanding the differences and interactions among these blocks is important.
In this paper, we uncover a clear {\bf sharpness disparity} across these blocks, which emerges early in training and intriguingly persists throughout the training process. Motivated by this finding, we propose {\bf Blockwise Learning Rate (LR)}, a strategy that tailors the LR to each block’s sharpness, accelerating large language model (LLM) pre-training.
By integrating Blockwise LR into AdamW, we consistently achieve lower terminal loss and nearly $2\times$ speedup compared to vanilla AdamW. We demonstrate this acceleration  across GPT-2 and LLaMA, with model sizes ranging from 0.12B to 2B and datasets of OpenWebText, MiniPile, and C4.
Finally, we incorporate Blockwise LR into other optimizers such as Adam-mini~\citep{zhang2024adam}, a recently proposed memory-efficient variant of Adam, achieving a combined $2\times$ speedup and $2\times$ memory saving. These results underscore the potential of exploiting the sharpness disparity to improve LLM training.
\end{abstract}

\doparttoc 
\faketableofcontents 
\part{} 
\vspace*{-4em}

\vspace{-.4cm}

\section{Introduction}
\label{section: introduction}

Transformers~\cite{vaswani2017attention} have achieved remarkable success across fields, including natural language processing~\cite{brown2020language}, vision~\cite{dosovitskiy2020image}, and scientific computing~\cite{jumper2021highly}. They have become the de facto design in modern AI models~\cite{team2023gemini,achiam2023gpt,liu2024deepseek}.

Compared to traditional architectures, e.g., multilayer perceptrons (MLPs), convolutional neural networks (CNNs), and recurrent neural networks (RNNs), transformers exhibit  distinctive {\bf alloy-like characteristics}, where {\bf diverse types of blocks} synergistically combine to achieve superior performance. A transformer at minimum includes self-attention (further broken down into query-key (\QK) and value-output (\VO)) blocks, point-wise feedforward networks (\FFN), normalization layers (\LN), and embedding layers (\Embed).  
Uncovering the distinct properties of these blocks, as well as the differences and interactions among them, is thus crucial for gaining a deeper insight into transformer models~\citep{wang2024understanding}.

In practice, transformers are typically trained using the AdamW optimizer~\cite{kingma2014adam,loshchilov2017decoupled}. Dissecting the alloy-like characteristics of transformers can provide insights into why Adam outperforms  stochastic gradient descent (SGD) for transformer training~\cite{devlin2018bert,zhang2020adaptive,pesme2023saddle,kunstner2024heavy,zhang2024transformers} and even holds promise for unlocking further improvements in training efficiency~\cite{popel2018training,xiong2020layer,zhang2024adam}. Particularly, \citet{zhang2024transformers} and \citet{zhang2024adam}  observed that unlike MLPs and CNNs,  the Hessian (aka sharpness or curvature) of transformers exhibits a distinct blockwise heterogeneity. Building on this insight, \citet{zhang2024adam} successfully reduced Adam's memory footprint nearly by half  without sacrificing training efficiency for  a variety of LLM and non-LLM training tasks.

\begin{figure}[!ht]
    \centering
    \includegraphics[width=0.42\linewidth]{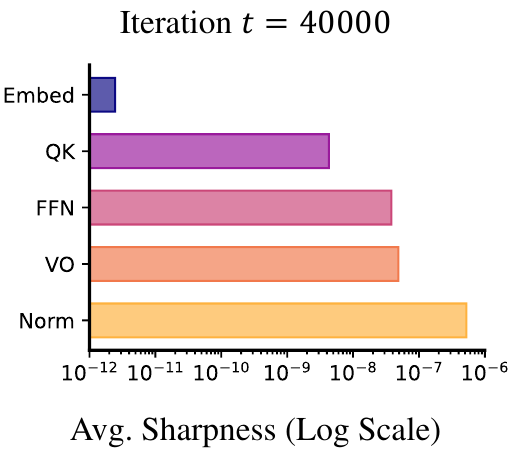}
    \includegraphics[width=0.55\linewidth]{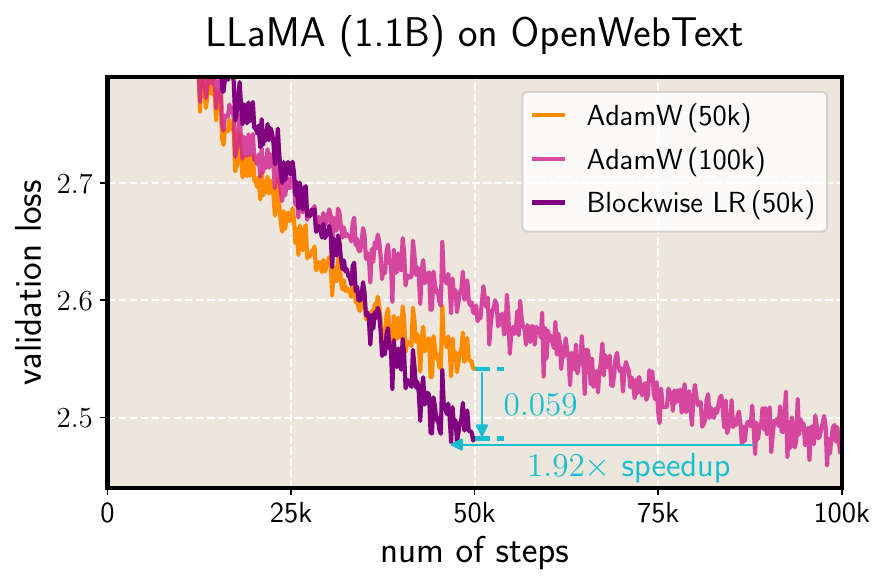}
    \vspace{-.1cm}
    \caption{(left) Sharpness disparity among block types in a pre-trained GPT-2 (small) on OpenWebText, exhibiting a clear order relationship as characterized by {\bf Principle}~\eqref{equ: main findings}.
    (right) For the pre-training of LLaMA (1.1B) on OpenWebText, incorporating our Blockwise LR strategy into AdamW results in a lower terminal loss and a $1.92\times$ speedup compared to the well-tuned vanilla AdamW.}
    \label{fig: introduction}
    \vspace{-.3cm}
\end{figure}

\paragraph*{Our Contribution.}
In this work, we aim to explore how we can leverage the aforementioned alloy-like characteristics of transformers to improve training efficiency. Specifically, our contributions can be summarized as follows:
\begin{itemize}[leftmargin=*, topsep=0pt]
\item 
\textbf{The sharpness disparity principle}. 
Motivated by the  alloy-like characteristics,  we examine the sharpness of transformers at the level of block type. 
Surprisingly, we discover a distinct disparity in sharpness across different block types, summarized as follows:
\begin{snugshade}
\begin{center}
\vspace*{-1.3em}
\begin{equation}\label{equ: main findings}
\hspace*{-.6em}\cS(\text{\Embed})\! \ll\!  \cS(\text{\QK})\!  <\!  \cS(\text{\FFN}) \! < \! \cS(\text{\VO})\!  \ll\!  \cS(\text{\Norm})\!
\end{equation}
\end{center}
\end{snugshade}


Here $\cS(\bullet)$ denotes the average sharpness of block type $\bullet$ (see Eq.\eqref{equ: mean sharpness} for the calculation details). 
See Figure~\ref{fig: introduction}~(left) for an illustration of this principle. Intriguingly, this principle emerges in the early training stage and persists throughout the subsequent training process, as shown in Figure~\ref{fig: law gpt process}. These findings are validated  through extensive experiments on the training of GPT-2~\citep{radford2019language} and LLaMA models~\citep{touvron2023LLaMA}, spanning various model sizes and datasets. We also provide preliminary theoretical explanations to complement these empirical observations.

\item \textbf{The Blockwise LR strategy.~} Inspired by \textbf{Principle}~\eqref{equ: main findings}, we propose tuning LRs by block type to accelerate LLM  pre-training. 
Specifically, we adjust the LRs of blocks within the same type in proportion to their sharpness, while keeping the LR of the block type with the highest sharpness unchanged.
This strategy accelerates the dynamics along low-sharpness directions without compromising training stability, as the latter  is governed by the high-sharpness directions.

The effectiveness of Blockwise LR is extensively validated in LLM pre-training across both GPT-2 and LLaMA models, with model sizes ranging from 0.12B to 2B parameters, and datasets including OpenWebText~\citep{Gokaslan2019OpenWeb}, MiniPile~\citep{kaddour2023minipile}, and C4~\citep{raffel2020exploring}. The results can be summarized as follows:
\begin{snugshade}
\begin{center}
\vspace*{.1em}
AdamW with Blockwise LR 
achieves lower terminal loss and is nearly {\bf $2\times$ faster} than vanilla AdamW.
\end{center}
\vspace{-0.8em}
\end{snugshade}
\vspace{-0.4em}
See Figure~\ref{fig: introduction} (right) for a quick view of the acceleration effect achieved by Blockwise LR.
Furthermore, we explore the compatibility of Blockwise LR with other Adam-based optimizers. Specifically, we integrate our Blockwise LR into Adam-mini~\citep{zhang2024adam}, achieving both $2\times$ speedup and $2\times$ memory saving. 

\end{itemize}




\vspace*{.4em}
\begin{remark}
There has been a long-standing effort in deep learning to accelerate neural network training by adapting layerwise learning rates, a strategy that has proven effective in architectures such as MLPs and CNNs~\citep{yang2019xlnet, yang2022tensor, everett2024scaling, shin2024initializing}. However, these approaches have not been successfully transferred to the training of deep transformers. We hypothesize that 
this gap stems from transformers' distinctive alloy-like characteristics: the inherent block-level diversity makes layerwise 
 learning rate strategies inadequate. To investigate this further, we examine layer-level sharpness in Figure~\ref{fig: layerwise no law} and no clear trends emerge across layers. This suggests that while sharpness disparity exists at the block-type level, it does not exhibit a consistent pattern at the layer level.
\end{remark}


\section{Related Works}
\label{section: related works}



{\bf Sharpness structures in transformers.} Recent work has started to investigate blockwise sharpness patterns in transformer models through Hessian-based analyses.
For example,~\citet{zhang2024transformers} empirically observed the sharpness' blockwise heterogeneity  but did not establish a clear principle regarding the sharpness disparity among different blocks. Meanwhile, \citet{ormaniec2024does} provided a Hessian analysis for a single self-attention (\SA) layer, focusing only on the  sharpness disparity between the query-key (\QK) and value-output (\VO) blocks within the same layer. 


In contrast, we examine sharpness at the block-type level across the entire transformer architecture, rather than focusing on individual blocks (as in \citet{zhang2024transformers}) or a single layer (as in \citet{ormaniec2024does}). This coarse-grained perspective reveals a consistent  disparity, as formalized by \textbf{Principle}~\eqref{equ: main findings},  which persists throughout most of the training process—except during the initial steps.


{\bf Efficient optimizers for LLM pre-training.} AdamW (Adam with decoupled weight decay)~\cite{loshchilov2017decoupled} has become the default optimizer in LLM pre-training. 
Efforts to design more efficient optimizers generally fall into two main categories: accelerating convergence and reducing memory footprint.
Accelerations have been developed using techniques such as Nesterov momentum~\citep{xie2022adan}, diagonal second-order estimates~\citep{liu2023sophia,wang2024improving}, variance reduction~\citep{yuan2024mars}, and matrix-based preconditioners~\citep{jordan2024muon,vyas2024soap}. 
Memory-efficient optimizers utilize sign-based methods~\citep{chen2024symbolic}, reduced usage of second moments in Adam~\citep{zhang2024adam}, and gradient low-rank projection~\citep{zhao2024galore}.
The closest work to our Blockwise LR is~\citet{wang2024improving}, which also increases the LR along low-sharpness directions. A detailed comparison is deferred to Section~\ref{section: application}.


{\bf The edge of stability (EoS) phenomenon.} Neural network training typically occurs at the EoS stage~\citep{wu2018sgd,Jastrzebski2020The,cohen2021gradient,cohen2022adaptive}, where the optimizer exhibits oscillatory behavior along sharp directions without diverging, while steadily progressing along flat directions, leading to loss reduction. Several works \citep{wen2024understanding,song2024does,cohen2024understanding,wang2024improving} have highlighted the crucial role of the dynamics along flat directions (referred to as river directions by \citet{wen2024understanding}, bulk directions by \citet{song2024does}, and stable direction in \citet{wang2024improving}) in reducing total loss. Notably, \citet{wen2024understanding} further demonstrated that this picture is essential for understanding LLM pre-training. Building on these insights, our Blockwise LR approach is designed to accelerate training by amplifying the dynamics particularly along the flat river directions.



\section{Preliminaries}


{\bf Notations.} Let $\norm{\cdot}_2$, $\norm{\cdot}_\rF$, and $\Tr(\cdot)$ denote the spectral norm, Frobenius norm and trace for matrices, respectively. Given $\bA\in\bbR^{m\times n}$, its row-wise vectorization is defined as $\vec(\bA)=(a_{1,1},\cdots,a_{1,n},\cdots,a_{m,1},\cdots,a_{m,n})\in\bbR^{mn}$. 
The Kronecker product and Hadamard product are denoted by $\otimes$ and $\odot$, respectively.
The row-wise mean and covariance of $\bA \in \bbR^{m \times n}$ are denoted by $\bbE_r[\bA]\in\bbR^{m\times n}$ and $\bbV_r[\bA]\in\bbR^{m\times n}$, respectively. Specifically, they are defined as: for all $i\in [m], j\in [n]$, 
$(\bbE_r[A])_{i,j}\!=\!\frac{1}{n}\sum_{k=1}^n A_{i,k},\, (\bbV_r[A])_{i,j}\!=\!\left(A_{i,j}-\frac{1}{n}\sum_{k=1}^n A_{i,k}\right)^2$.
We will use standard big-O notations like $\cO(\cdot)$, $\Omega(\cdot)$, and $\Theta(\cdot)$ to hide problem-independent constants.

{\bf Jacobian matrix.} Given a vector-valued function: $\boldsymbol{b}\mapsto\ba(\boldsymbol{b})$ with  $\boldsymbol{b}\in\bbR^n$ and $\ba(\boldsymbol{b})\in\bbR^{m}$, the Jacobian is defined as $\frac{\partial \ba}{\partial\boldsymbol{b}}=(\frac{\partial a_i}{\partial b_j})_{i,j}\in\bbR^{m\times n}$. Analogously, for a matrix-valued function: $\bB\mapsto\bA(\bB)$ where
$\bB\in\bbR^{p \times q}$ and $\bA(\bB)\in\bbR^{m\times n}$, to avoid directly working with tensors, the Jacobian is defined as $\frac{\partial \bA}{\partial\bB}:=\frac{\partial\vec(\bA)}{\partial\vec(\bB)}\in\bbR^{mn\times pq}$.


\subsection{The Transformer Architecture}

\vspace{-.1cm}

Given an $n$-token input sequence $\bX=(\bx_{1}^\top,\cdots,\bx_{n}^\top)^\top\in\bbR^{n\times d}$, where $d$ refers to the vocabulary size in LLM and each $\bx_i$ corresponds to the token's one-hot encoding,  an $L$-layer transformer $\TF$ processes it as follows.

{\bf Embedding layer.} First, each input token is embedded into the latent space through an embedding layer with parameters $\bW_E\in\bbR^{d\times D},\boldsymbol{b}_E\in\bbR^{1\times D}$:

\vspace{-.5cm}

$$
\bx_s^{(0)}=\bx_{s}\bW_E+\boldsymbol{b}_E,\ s\in[n],
$$ 

\vspace{-.2cm}

where the bias $\boldsymbol{b}_E$ is omitted in LLMs such as nanoGPT~\citep{Karpathy2022}.

{\bf $L$-layer \SA-\FFN\ blocks.}
Then the embedded sequence $\bX^{(0)}$ is processed by  $L$-layer \SA-\FFN\ blocks, and the output of the final layer is taken as the output sequence $\TF(X)=X^{(L)}\in\bbR^{n\times D}$. For each layer $l\in[L]$, the computations are as follows:

\vspace{-.6cm}

\begin{equation}\label{model: Transformer}
\begin{aligned}
    \bX^{(l-\frac{1}{2})}&=\bX^{(l-1)}+\SA^{(l)}(\LN^{(l-1/2)}(\bX^{(l-1)}));
    \\
    \bX^{(l)}&=\bX^{(l-\frac{1}{2})}+\FFN^{(l)}(\LN^{(l)}(\bX^{(l-\frac{1}{2})})).
\end{aligned}
\end{equation}

\vspace{-.3cm}

{\bf \LN\ blocks.}
Here, $\LN^{(v)}$ ($v\in\{l-1/2,l\}$) denote normalization layers (e.g., LayerNorm~\cite{lei2016layer} and RMSNorm~\citep{zhang2019root}) with learnable parameters $\bgamma^{(v)},\bbeta^{(v)}\in\bbR^{1\times D}$.
For LayerNorm, the computation for a token $\bx\in\bbR^{1\times D}$ is:

\vspace{-.6cm}

\begin{align*}
    \LN^{(v)}(\bx)=\frac{\bx-\bbE_r[\bx]}{\bbV_r[\bx]}\odot\bgamma^{(v)}+\bbeta^{(v)}.
\end{align*}

\vspace{-.3cm}

where the bias $\bbeta$ is omitted in LLMs such as nanoGPT. 

{\bf \FFN\ blocks.}
$\FFN^{(l)}$ denotes a (token-wise) two-layer FFN of width $M$, comprising parameters $\bW_1^{(l)}\in\bbR^{D\times M},\bW_2^{(l)}\in\bbR^{M\times D}$, and using activation function $\sigma(\cdot)$ such as ReLU. For any token $\bx\in\bbR^{1\times D}$, the operation is:

\vspace{-.6cm}

\begin{align*}
    \FFN^{(l)}(\bx)=\sigma(\bx\bW_1^{(l)})\bW_2^{(l)}.
\end{align*}

\vspace{-.3cm}

{\bf \SA\ blocks.} $\SA^{(l)}$, a multi-head self-attention, has parameters $\bW_{Q}^{(l)},\bW_{K}^{(l)},\bW_{V}^{(l)},\bW_O^{(l)}\in\bbR^{D\times D}$. When applied to a sequence $\bZ\in\bbR^{n\times D}$, it operates as:

\vspace{-.6cm}

\begin{gather*}
    \SA^{(l)}(\bZ)=\sum_{h=1}^{H}\SA^{(l,h)}(\bZ)\bW_O^{(l,h)},\quad \SA^{(l,h)}(\bZ)=
    \\\sm\left(\frac{\<\bZ\bW_Q^{(l,h)},\bZ\bW_K^{(l,h)}\>+\bM}{\sqrt{D/H}}\right)\left(\bZ\bW_V^{(l,h)}\right),
\end{gather*}

\vspace{-.3cm}

where $H$ is the head number, and $\bW_{Q}^{(l,h)},\bW_{K}^{(l,h)},\bW_{V}^{(l,h)}\in\bbR^{D\times(D/H)}$, $\bW_{O}^{(l,h)}\in\bbR^{(D/H)\times D}$ are split from $\bW_{Q}^{(l)},\bW_{K}^{(l)},\bW_{V}^{(l)}$, $\bW_O^{(l)}$ by heads, respectively. The operator
$\sm(\cdot)$ represents the row-wise softmax normalization. 
For the next-token prediction, the mask $\bM\in\bbR^{n\times n}$ satisfies $M_{i,j}=-\infty$ if $i< j$ and $M_{i,j}=0$ otherwise.



\subsection{Blockwise  Sharpness and the Efficient Estimation} 

\vspace{-.1cm}

Measuring sharpness requires accessing the Hessian matrix, which is computationally expensive due to the high dimensionality of the parameter space. Consequently, approximate methods are needed to reduce computational complexity.

Let $\ell(\cdot,\cdot)$ denote the cross-entropy loss. For an input data $\bx\in\bbR^{d_x}$ and label $\by\in\bbR^{d_y}$, let the model's prediction be $f(\bx;\btheta)\in\bbR^{d_y}$.
The Fisher (Gauss-Newton) matrix $F(\btheta)$ is widely recognized approximation of the Hessian, particularly near minima. Thus, the diagonal Hessian can be estimated as $\bh(\btheta)={\rm diag}(F(\btheta))$, a popular technique in deep learning optimization~\citep{martens2015optimizing,grosse2016kronecker,george2018fast,mi2022make,liu2023sophia,wang2024improving}. Moreover,
given an input batch $\{(\bx_b,\by_b)\}_{b=1}^B$, the empirical diagonal Fisher can be estimated:
$
{\rm diag}(\hat{F}(\btheta))=\frac{1}{B}\sum_{b=1}^B\nabla\ell(f(\bx_b;\btheta);\hat{\by}_b)\odot\nabla\ell(f(\bx_b;\btheta);\hat{\by}_b), \text{ where }\hat{\by}_b\sim{\rm softmax}(f(\btheta;\bx_b)).
$ 
However, as noted by~\citet{liu2023sophia},
implementing this estimator is computationally expensive due to the need to calculate $B$ single-batch gradients.
\citet{liu2023sophia} proposed a more convenient estimator ${\rm diag}(\hat{F}_{\rm eff}(\btheta))$, which only requires the computation of the mini-batch gradient $\nabla\hat{\cL}_B(\btheta)=\frac{1}{B}\sum_{b=1}^B\nabla\ell(f(\bx_b;\btheta);\hat{\by}_b) \text{ with }\hat{\by}_b\sim{\rm softmax}(f(\bx_b;\btheta))$:
\begin{equation}\label{equ: fisher estimate}
\begin{aligned}
   \bh(\btheta) = {\rm diag}(\hat{F}_{\rm eff}(\btheta))=B\cdot\nabla\hat{\cL}_B(\btheta)\odot\nabla\hat{\cL}_B(\btheta).
\end{aligned}
\end{equation}
According to~\citet[Section 2]{liu2023sophia}, this estimator is unbiased, i.e., $\bbE_{\hat{\by}}[{\rm diag}(\hat{F}_{\rm eff}(\btheta))]=\bbE_{\hat{\by}}[{\rm diag}(\hat{F}(\btheta))]$. 

Given a block type $\bullet\in\{\Embed,\QK,\VO,\FFN,\LN\}$, let $\btheta[\bullet]$ represent the parameters associated with all blocks of that type,  and let $\bh(\btheta[\bullet])$ denote the corresponding diagonal Hessian. The average sharpness for each block type can then be approximated as follows:

\vspace{-.6cm}

\begin{equation}\label{equ: mean sharpness}
\cS(\btheta[\bullet]):=\frac{\Tr(\bh(\btheta[\bullet]))}{\#(\btheta[\bullet])}= \frac{B\norm{\nabla_{\btheta[\bullet]}\hat{\cL}_B(\btheta)}_\rF^2}{\#(\btheta[\bullet])},    
\end{equation}

\vspace{-.2cm}

where $\hat{\cL}_B$ corresponds to~\eqref{equ: fisher estimate} and $\#(\btheta[\bullet])$ denotes the number of parameters associated with the block type $\bullet$. For brevity, $\btheta$ in~\eqref{equ: mean sharpness} will be omitted when there is no ambiguity. 
\begin{remark}
It is worth noting that in~\eqref{equ: mean sharpness}, the sharpness is averaged over all blocks of the same type, which may be distributed across different layers, rather than being calculated within each individual block.
\end{remark}


\section{The Sharpness Disparity Principle}\label{section: principle}


\subsection{Main Findings}


We first  investigate the sharpness  disparity across different types of  building blocks (\Embed, \QK, \VO, \FFN, \Norm) in transformer-based LLMs. 
Specifically, we pre-trained GPT-2~\citep{radford2019language} and LLaMA~\citep{touvron2023LLaMA} models on the OpenWebText dataset using default configurations.
Blockwise diagonal Hessians are analyzed at various checkpoints using the Hessian estimator~\eqref{equ: fisher estimate}. The experimental details can be found in  Appendix~\ref{appendix: experiments for baselines}.

\begin{figure}[!htb]
    \centering
    \includegraphics[width=0.48\linewidth]{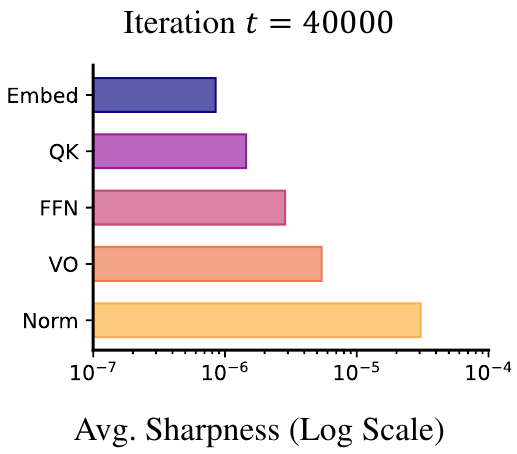}
    \includegraphics[width=0.45\linewidth]{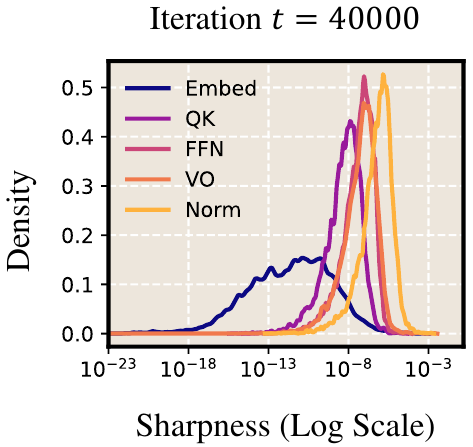}
    \vspace{-.1cm}
    \caption{(left) The average sharpness for the five typical block types in a pre-trained LLaMA model (0.25B); (right) the sharpness distribution across different blocks in a pre-trained GPT-2 (small) model.}
    \label{fig: law final}
    \vspace{-.3cm}
\end{figure}

In Figures~\ref{fig: introduction} (left) and \ref{fig: law final} (left), we report the average  sharpness, estimated using~\eqref{equ: mean sharpness}, of  the five typical types of blocks for GPT-2 and LLaMA, respectively. The results reveal a clear and consistent sharpness disparity among different block types, as summarized in  {\bf Principle}~\eqref{equ: main findings}. Specifically, \Norm\ layers consistently exhibit the highest sharpness, the \Embed\ layers are the flattest, and \QK\ layers are relatively flatter compared to \FFN\ and \VO\ layers.
These findings, to the best of our knowledge, provide the first comprehensive comparison of sharpness across  block types in transformers.

Figure~\ref{fig: law final} (right) plots the full sharpness distribution for each block type, whereas Figures~\ref{fig: introduction} (left) and~\ref{fig: law final} (left) only report mean sharpness values. Evidently,  even at the distribution level,  {\bf Principle} \eqref{equ: main findings} remains valid. Interestingly,
 the \Embed~block exhibits much higher variance compared to other blocks. This behavior likely stems from the embedding layer's direct interaction with the entire vocabulary, where rare tokens result in the wide spread of small sharpness and frequent tokens contribute to large sharpness. A similar insight has been utilized by  \citet{kunstner2024heavy} to explain the necessity of Adam in training NLP models.

\begin{figure*}[tb!]
    \centering
    \subfloat[Evolution of the average sharpness across different blocks during pre-training GPT-2 (small) on OpenWebText.]
    {\includegraphics[width=0.95\linewidth]{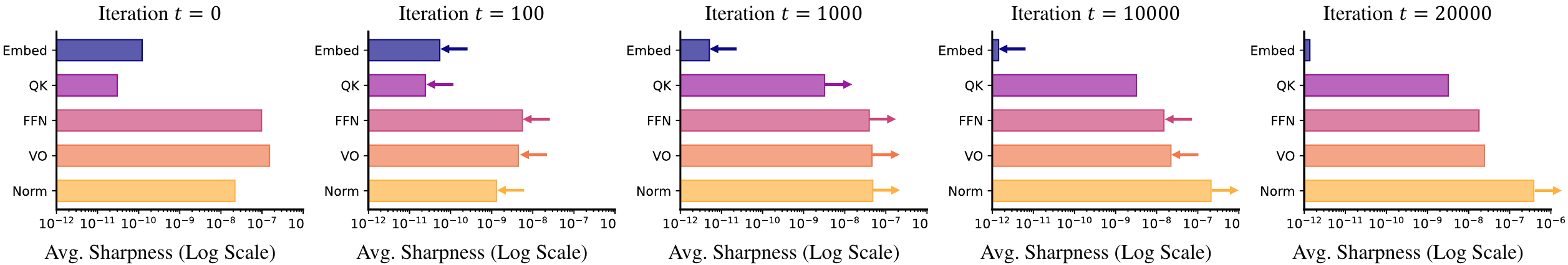} }
    \\
    \vspace{-.2cm}
    \subfloat[Evolution of the average sharpness across different blocks during pre-training LLaMA (0.25B) on OpenWebText.]
    {\includegraphics[width=0.95\linewidth]{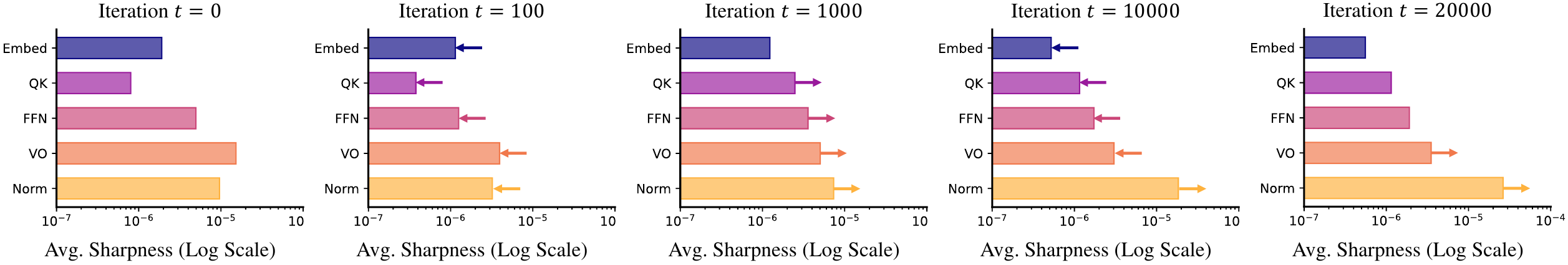} }
    \vspace{-.15cm}
    \caption{In these experiments, the total training steps are both 50k. {\bf Principle} \eqref{equ: main findings} emerges during the initial phase (from iteration 0 to iteration 1k), which accounts for only approximately $2\%$ of the total steps, and persists throughout the subsequent  training process.}
    \label{fig: law gpt process}
    \vspace{-.3cm}
\end{figure*}

Furthermore, Figure~\ref{fig: law gpt process} illustrates the  evolution of blockwise sharpness during the training process. We can see that {\bf Principle} \eqref{equ: main findings} is not exclusive to well-trained transformers; instead, it emerges in the early stages of training and persists consistently throughout the subsequent training process. This observation underscores the potential of leveraging {\bf Principle}~\eqref{equ: main findings} to enhance LLM pre-training; we refer to Section~\ref{sec:adam_bl_adam_with_blockwise_learning_rates} for further explorations.

{\bf Comparison with existing works}. Our findings build on prior work, extending key observations. \citet{zhang2024transformers}  noted the block heterogeneity in the Hessian of transformers  but did not establish a clear principle for sharpness distinctions across blocks, as we do with {\bf Principle}~\eqref{equ: main findings}. The work of \citet{ormaniec2024does} is more closely related  but focuses solely on a single self-attention layer (\SA), reporting the relationship $\cS(\QK) < \cS(\VO)$. In contrast, we analyze all major block types in transformers, including \Embed, \FFN, and \LN, thereby offering a more comprehensive principle that captures the full scope of  sharpness disparity.


\subsection{Theoretical Insights}
\label{section: theoretical insights}


To provide theoretical insights into explaining {\bf Principle}~\eqref{equ: main findings}, we derive analytic expressions of $\cS(\bullet)$ and analyze their dependence on parameter magnitudes and numbers of each block. For simplicity, we denote $\cQ(\btheta):=\hat{\cL}_B(\btheta)$, where $\hat{\cL}_B(\btheta)$ is defined in~\eqref{equ: fisher estimate}. Then from~\eqref{equ: mean sharpness}, we have $\cS(\bullet)=B\norm{\nabla_{\bullet}\cQ}_{\rF}^2/\#(\bullet)$. Without loss of generality, we set $B=1$.
Our calculations for $\nabla \cQ$ apply to general $\cQ$. 

Considering blocks across different layers is  complicated. Therefore, we focus on comparisons within the same layer. Specifically, we examine the following sharpness comparisons: (i) \FFN\ vs. \LN\ within the same layer; (ii) \SA\ (comprising \QK\ and \VO) vs.~\LN\ within the same layer; and (iii) \Embed\ vs.~the adjacent \LN.

\begin{theorem}[\FFN\ vs.~\Norm]\label{thm: FFN vs Norm}
Consider the $l$-th layer in a transformer~\eqref{model: Transformer}.
Omitting the layer index for simplicity, let $\bY=\bX+\FFN\left(\LN\left(\bX;\bgamma\right);\bW_1,\bW_2\right)$, where FFN utilizes the (Leaky) ReLU activation $\sigma$. 
Then, the gradients of $\cQ$ w.r.t. $\bW_1,\bW_2$, and $\bgamma$ are:
\begin{align*}
    \frac{\partial \cQ}{\partial\bW_{2}}&=\frac{\partial \cQ}{\partial \bY}\left(\bX_{\LN}\bW_1\odot\frac{\partial\ASA}{\partial\MSA}\right)\otimes\bI_d;
    \\
    \frac{\partial \cQ}{\partial\bW_{1}}&=\frac{\partial \cQ}{\partial \bY}\left(\bI_{n}\otimes{\bW_2}^\top\right)\frac{\partial\ASA}{\partial\MSA}\left(\bX_{\LN}\otimes\bI_M\right);
    \\
    \frac{\partial \cQ}{\partial\bgamma}&=\frac{\partial \cQ}{\partial \bY}\left(\bI_{n}\otimes{\bW_2}^\top\right)\frac{\partial\ASA}{\partial\MSA}\left(\bI_n\otimes{\bW_1}^\top\right)
    \\&\quad\quad\quad \diag\big(\vec(\bX_{\rm std})\big)\big(\bone_{n\times 1}\otimes \bI_{d}\big),
\end{align*}  
where $\bX_{\rm std}:=\frac{\bX-\bbE_r[\bX]}{\sqrt{\bbV_r[\bX]}},\bX_\LN:=\LN(\bX;\bgamma)=\bX_{\rm std}\odot\big(\bone_{n\times 1}\otimes\bgamma\big),\ASA:=\sigma(\MSA),\MSA:=\bX_{\LN}\bW_1$. 
Let $\Psi:=n\sqrt{D}\norm{\frac{\partial \cQ}{\partial \bY}}_{\rF}\norm{\frac{\partial\ASA}{\partial\MSA}}_{\rF}\norm{\bW_1}_\rF\norm{\bW_2}_\rF\norm{\bgamma}_{\rF}$. 
Then, the blockwise average sharpness can be bounded as: 

\vspace{-.6cm}

\begin{align*}
    \cS(\bW_{\bullet})&=\cO\left(\frac{\Psi^2}{D^2\|\bW_\bullet\|_\rF^2}\right),\bullet\in\{1,2\};\\ 
    \cS(\bgamma)&=\cO\left(\frac{\Psi^2}{D\|\bgamma\|_\rF^2}\right),
\end{align*}

\vspace{-.3cm}

where the denominators ($D^2$ or $D$) reflect the number of parameters in each group.
\end{theorem}

\vspace{-.1cm}

Theorem~\ref{thm: FFN vs Norm} provides theoretical support for our main finding: $\cS(\FFN)$ is substantially smaller than $\cS(\Norm)$.
As illustrated in Figure~\ref{fig: blockwise norm} (a), during training, $\norm{\bgamma}_\rF$ gradually decreases, and $\norm{\bW_{\bullet}}_\rF$ ($\bullet\in\{1,2\}$) in FFN layers remains larger than $\norm{\bgamma}_{\rF}$, resulting in $D^2\norm{\bW_\bullet}_{\rF}^2\gg D\norm{\bgamma}_{\rF}^2$.

\begin{theorem}[\QK, \VO\ vs.~\Norm]\label{thm: QK VO vs Norm}
Consider the ($l-\frac{1}{2}$)-th layer in~\eqref{model: Transformer}. Omitting the layer index for simplicity, let $\bY=\bX+\SA\Big(\LN\left(\bX;\bgamma\right);\bW_K,\bW_Q,\bW_V,\bW_O\Big)$. 
Consider a single-head attention (i.e., $H=1$) for simplicity. Then, the gradients of $\cQ$ w.r.t. different blocks ($\bW_K,\bW_Q, \bW_V,\bW_O, \bgamma$) are provided in Appendix~\ref{appendix: proof: thm: QK VO vs Norm}.
Furthermore, there exist two problem-dependent constants $\Phi,\Psi>0$ (detailed in Appendix~\ref{appendix: proof: thm: QK VO vs Norm}), such that:

\vspace{-.6cm}

\begin{align*}
    \cS(\bW_{\bullet})&=\cO\left(\frac{\Phi^2}{D^2\norm{\bW_\bullet}_\rF^2}\right),\ \bullet\in\{K,Q\};
    \\
    \cS(\bW_{\bullet})&=\cO\left(\frac{\Psi^2}{D^2\norm{\bW_\bullet}_\rF^2}\right),\ \bullet\in\{V,O\};
    \\
    \cS(\bgamma)&=\cO\left(\frac{\Phi^2 + \Psi^2}{D\norm{\bgamma}_\rF^2}\right).
\end{align*}

\vspace{-.4cm}

where the denominators ($D^2$ or $D$) reflect the number of parameters in each group.
\end{theorem}

\vspace{-.1cm}

Theorem~\ref{thm: QK VO vs Norm} provides theoretical support for our main finding that both $\cS(\QK)$ and $\cS(\VO)$ are significantly smaller than $\cS(\LN)$.
The inclusion of the {\rm softmax} operation in attention layers introduces additional complexity in the calculations. Detailed derivations are given in the appendix. As shown in Figure~\ref{fig: blockwise norm} (b), during training, $\norm{\bgamma}_\rF$ gradually decreases, and $\norm{\bW_{\bullet}}_\rF$ ($\bullet\in\{K,Q,V,O\}$) in \SA\ blocks remains larger than $\norm{\bgamma}_{\rF}$, resulting in $D^2\norm{\bW_\bullet}_{\rF}^2\gg D\norm{\bgamma}_{\rF}^2$.

This theorem does not explicitly establish that $\cS(\QK) < \cS(\VO)$. Studying this relation requires a deeper analysis of the constants $\Phi$ and $\Psi$, as well as the magnitudes of $\norm{\bW_\bullet}_\rF$. \citet{ormaniec2024does} has demonstrated $\cS(\QK)<\cS(\VO)$ both theoretically and experimentally, and we defer to that analysis instead of repeating it here.

\begin{theorem}[\Embed\ v.s.~\Norm]\label{thm: Embed vs Norm}
Consider the embedding layer and its adjoint normalization layer of a transformer~\eqref{model: Transformer}. Omitting the layer index for simplicity, let: $\bY:=\LN(\bX\bW_{\rm emb};\bgamma)$.
The gradients of $\cQ$ w.r.t~$\bW_{\rm emb}$ and $\bgamma$ are derived in Appendix~\ref{appendix: proof: thm: Embed vs Norm}.
Moreover, there exists a problem-dependent constant $\Psi>0$ (also detailed in Appendix~\ref{appendix: proof: thm: Embed vs Norm}), such that:

\vspace{-.6cm}

\begin{align*}
    \cS(\bW_E)&=\cO\Bigg(\frac{\Psi^2}{Dd\min\limits_{i\in[d]}\norm{\tilde{\bw}_{E_i}}_2^2}\Bigg);\\  
    \cS(\bgamma)&=\cO\left(\frac{\Psi^2}{D\norm{\bgamma}_\rF^2}\right),
\end{align*}

\vspace{-.4cm}

where $\tilde{\bW}_E=(\tilde{\bw}_{E_1}^\top,\cdots,\tilde{\bw}_{E_d}^\top)^\top:=\bW_E-\bbE_r[\bW_E]$. The denominators ($Dd$ or $D$) represent the number of parameters in each group.
\end{theorem}

\vspace{-.1cm}

Theorem~\ref{thm: Embed vs Norm} provides theoretical justification for our main finding that $\cS(\Embed)$ is much smaller than $\cS(\Norm)$. 
As shown in Figure~\ref{fig: blockwise norm}(c), during training, $Dd\norm{\tilde{\bw}_{E_i}}_2^2\gg D\norm{\bgamma}_F^2$. (Notice that the vocabulary size $d$ is very large in practice, e.g., 50304 for the GPT tokenizer.)

Recalling the definition of average sharpness~\eqref{equ: mean sharpness}, the key step in deriving Theorem~\ref{thm: FFN vs Norm} and~\ref{thm: QK VO vs Norm}, and~\ref{thm: Embed vs Norm} is establishing  $\|\nabla_{\bullet}\cQ\|=\cO(1/ \|\btheta[\bullet]\|)$.
This relationship is highly intuitive given the compound multiplicative nature of transformer blocks, where the norm of the derivatives is inversely proportional to the norm of associated parameters, even with weak non-linearities. For example, if $y=\prod_{i=1}^n x_i$ and $\cQ=\varphi(y)$, then $|\partial\cQ/ \partial x_i|=|\phi'(y) y/ x_i|\propto 1/|x_i|$ for all $i\in[n]$.


\section{The Blockwise LR Strategy}\label{section: application}
\label{sec:adam_bl_adam_with_blockwise_learning_rates}



Recalling Figure~\ref{fig: law gpt process}, the sharpness disparity across different blocks, as described in~\eqref{equ: main findings}, emerges early in training and persists until convergence. This insight can be leveraged to accelerate LLM pre-training, as elaborated later.

{\bf Fast-slow dynamics at EoS.} 
As discussed in Section~\ref{section: related works}, recent studies \citep{wen2024understanding,song2024does,wang2024improving} have highlighted the distinct roles of the dynamics along high- and low-sharpness directions during EoS. The main picture is summarized as follows:

\vspace{-.2cm}

\begin{itemize}[leftmargin=2em]
    \item {\bf Fast dynamics}: Along {\em high-sharpness directions}, the optimizer exhibits significant fluctuations without converging or diverging. These components of dynamics govern training stability, as further increasing the LR in these directions can lead to instability, while contributing little to loss reduction.

    \vspace{-.1cm}
    
    \item {\bf Slow dynamics}: Along {\em low-sharpness directions}, the optimizer progresses steadily, making the primary contribution to loss reduction, albeit at a slow rate.
\end{itemize}

\vspace{-.1cm}

Inspired by the above picture, a promising approach to accelerating training is as follows: given a base optimizer, increase the LRs along low-sharpness directions while keeping the LR of high-sharpness directions unchanged. This strategy aims to  speed up loss reduction without compromising training stability.

\citet{wang2024improving} has implemented this idea by adjusting the LR of each parameter based on its sharpness. However, this approach faces two key challenges: 1) it requires frequent diagonal Hessian estimation, which imposes significant computational and memory overhead;  2) sharpness estimates at the individual parameter level can be unreliable.

{\bf The Blockwise LR.}
 Unlike \citet{wang2024improving},  we  propose adjusting LRs at the block-type level, as our {\bf Principle}~\eqref{equ: main findings}   reveals a consistent sharpness disparity at this granularity.
Specifically, let $\eta_{\rm base}$ denote the LR for base optimizers such as AdamW, the LR for each block type is then adjusted as follows:

\vspace{-.2cm}

\begin{itemize}[leftmargin=2em]
    \item \Norm\ blocks (the sharpest directions): we still use the base LR, $\eta_\Norm=\eta_{\rm base}$, to keep training stability;

    \vspace{-.1cm}
    
    \item  Other blocks (low-sharpness directions): we adjust the LRs of these blocks by $\eta_{\bullet}\propto r(\bullet) \eta_{\rm base}$, where $\bullet\in\{\Embed,\QK,\FFN,\VO\}$, where $r(\bullet)$ denotes the adjusting ratio for the block type $\bullet$.
\end{itemize}

\vspace{-.1cm}

Naturally, we can set $r(\bullet)\propto \cS(\Norm)/\cS(\bullet)$. However,
in practice, we find that manually tuning $r(\bullet)$'s--involving only four hyperparameters--while following the qualitative trend described by   {\bf Principle}~\eqref{equ: main findings} is more effective. Further details are provided in Section~\ref{sec:experiments}.

It is also worth noting that due to its simplicity, Blockwise LR can be seamlessly integrated into modern LLM training frameworks such as Megatron~\citep{shoeybi2019megatron}.


\vspace{-.1cm}

\section{Experiments}
\label{sec:experiments}


{\bf Models and datasets.} We evaluate our proposed Blockwise LR in the pre-training of decoder-only LLMs across various  model types, model sizes, and datasets\footnote{The code is available at \texttt{\href{https://github.com/Wongboo/BlockwiseLearningRate}{https://github.com/}}
\texttt{\href{https://github.com/Wongboo/BlockwiseLearningRate}{Wongboo/BlockwiseLearningRate}}.}.
Specifically, we consider two widely-used LLMs: {\bf LLaMA} and {\bf GPT-2}; we experiment with model sizes ranging {\bf from 0.12B to 2B} parameters; the datasets includes OpenWebText~\citep{Gokaslan2019OpenWeb}~\footnote{An opensource recreation of the WebText corpus, widely used for LLM pre-training such as RoBERTa~\citep{liu2019roberta} and GPT-2.}, MiniPile~\citep{kaddour2023minipile}\footnote{A 6GB subset of the deduplicated Pile (825GB)~\citep{gao2020pile}}, and Colossal Clean Crawled Corpus (C4)~\citep{raffel2020exploring}\footnote{A large-scale public language datasets, widely used for LLM pre-training such as T5~\citep{raffel2020exploring}}, providing a highly diverse text corpus.  


{\bf Baselines.} 
As a baseline, we use the default AdamW  optimizer, configured with the hyperparameters $\beta_1=0.9,\beta_2=0.95$ and weight decay $\lambda=0.1$. To ensure training stability, gradient clipping is applied with $1.0$. These settings align with the training protocols used in nanoGPT and LLaMA models~\citep{touvron2023LLaMA}.
The LR strategy includes a linear warm-up phase followed by a cosine decay scheduler, capped at \texttt{lr\_max}. And the terminal LR \texttt{lr\_min} is set to \texttt{lr\_max}/20. For each experiment, we {\em first tune} the \texttt{lr\_max} to be optimal for AdamW, and the baselines are trained using these optimal \texttt{lr\_max}'s.
Details of the tuned \texttt{lr\_max} values can be found in Appendix~\ref{appendix: experiments for baselines}.

{\bf Adjusting ratio tuning and its transferability.}
To incorporate the Blockwise LR into AdamW, we simply use the \texttt{lr\_max} (tuned for vanilla AdamW) for \LN\ blocks. Then, we {\bf only tuned the four adjusting ratios in a single small-scale experiment} -- specifically the pre-training of LLaMA (0.25B) on Minipile -- following the rule:  $r(\bullet)$ is adjusted according to the trend of $\frac{\cS(\Norm)}{\cS(\bullet)}$,  guided by {\bf Principle}~\eqref{equ: main findings}. The tuned hyperparameters are: 
\vspace{-0.1cm}
{\colorlet{shadecolor}{blue!10}
\begin{snugshade}
\vspace{-0.15cm}
\begin{center}
\vspace{-0.35cm}
\begin{equation}\label{equ: tuned hyperparameters, blockwise lr}
    r(\Embed)=10, r(\QK)=8, r(\FFN)=6, r(\VO)=4.
\end{equation}
\end{center}
\vspace{-0.3cm}
\end{snugshade}
\vspace{-0.2cm}
}

Notably, the adjusting ratios are highly robust hyperparameters, as demonstrated in the following ways:

\vspace{-.2cm}

\begin{itemize}
    \item First, as shown in Figure~\ref{fig: robust llama web}, in the experiments for tuning the adjusting ratios, Blockwise LR demonstrates robustness to these hyperparameters, consistently accelerating pre-training across a range of $r(\bullet)$'s. The configuration in~\eqref{equ: tuned hyperparameters, blockwise lr} achieves the largest improvements among those tested. Notably, even with suboptimal ratios, Blockwise LR still delivers significant performance gains. Further details are provided in Appendix~\ref{appendix: experiments, blockwise LR}.

\vspace{-.1cm}
\item Second, the configuration in~\eqref{equ: tuned hyperparameters, blockwise lr}, tuned from a single experiment,  {\bf transfers perfectly} across all AdamW experiments conducted in this paper. Consequently, {\bf we adopt \eqref{equ: tuned hyperparameters, blockwise lr} as the default adjusting ratios for all AdamW experiments}. This robustness aligns with the consistency of {\bf Principle}~\eqref{equ: main findings}, which holds across GPT and LLaMA models, various model sizes, and datasets.

\vspace{-.1cm}

\end{itemize}

\vspace{-.1cm}

\subsection{Main Results}

\begin{figure*}[!ht]
    \centering
    \includegraphics[width=0.24\linewidth]{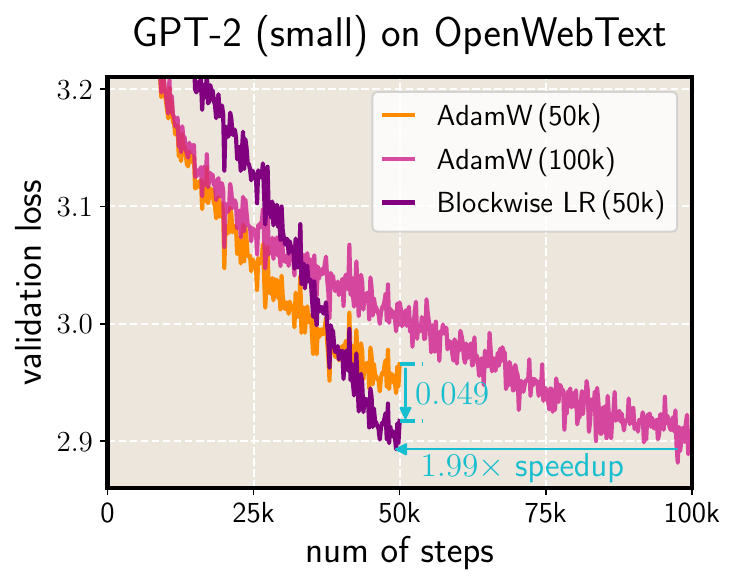}
    \includegraphics[width=0.24\linewidth]{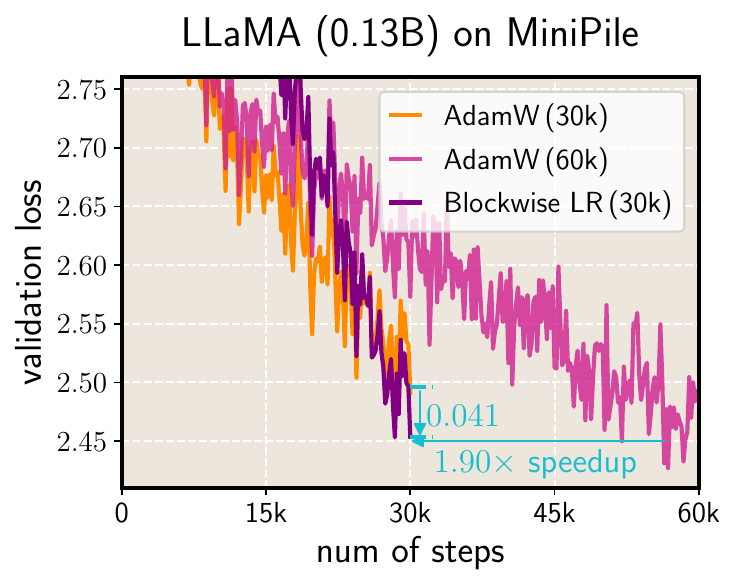}
    \includegraphics[width=0.24\linewidth]{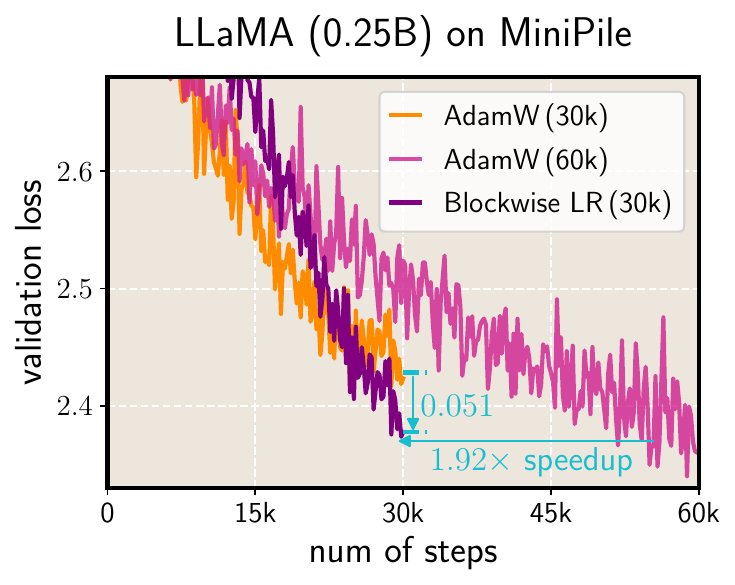}
    \includegraphics[width=0.24\linewidth]{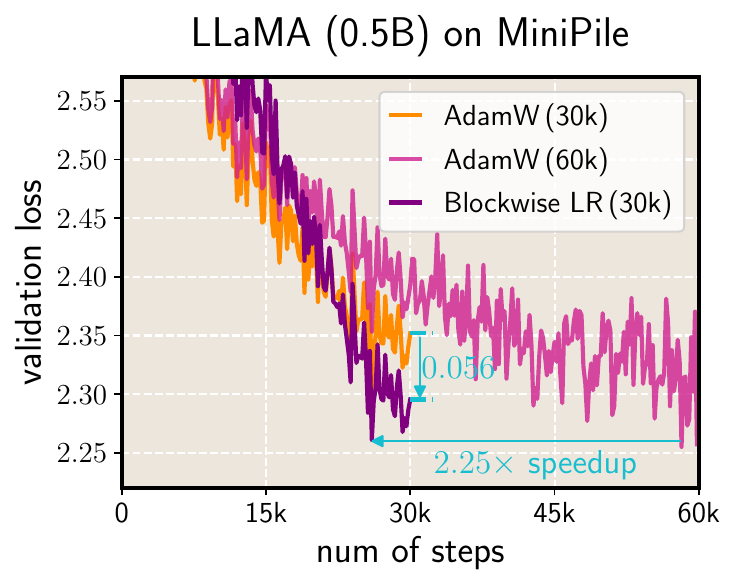}
    \includegraphics[width=0.24\linewidth]{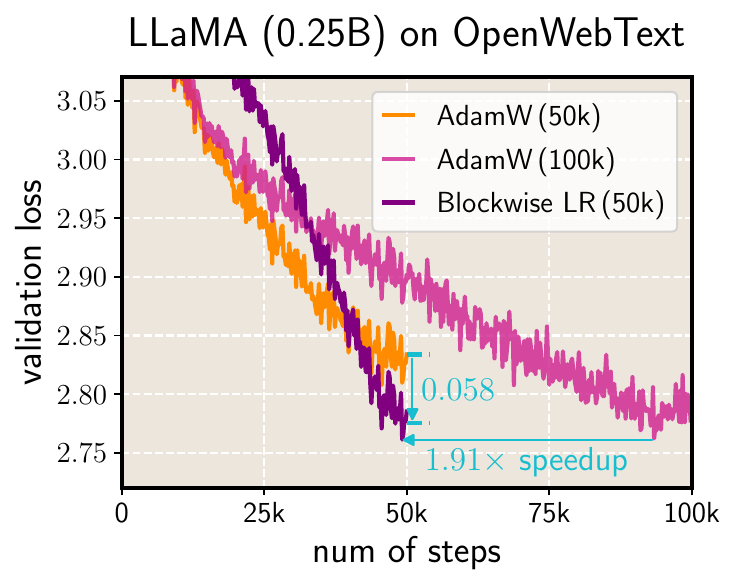}
    \includegraphics[width=0.24\linewidth]{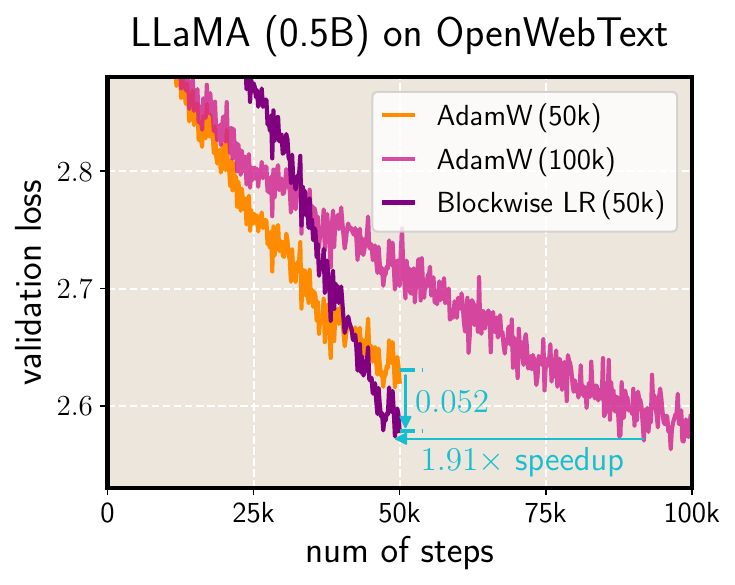}
    \includegraphics[width=0.24\linewidth]{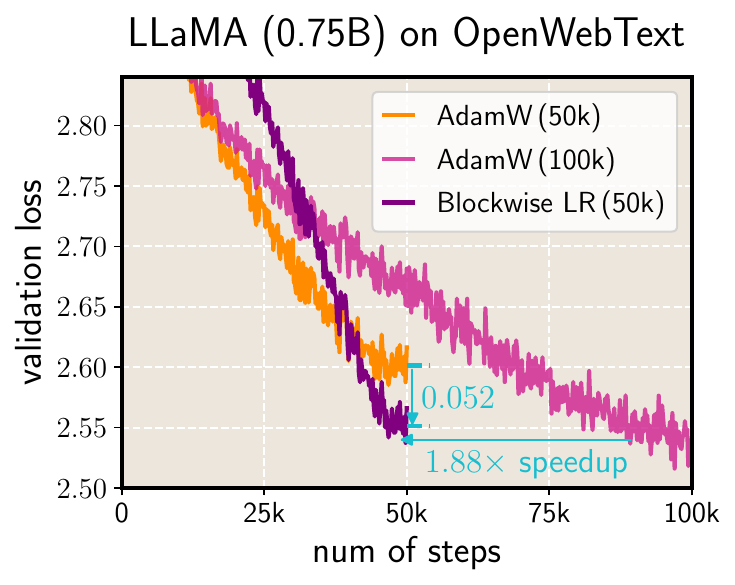}
    \includegraphics[width=0.24\linewidth]{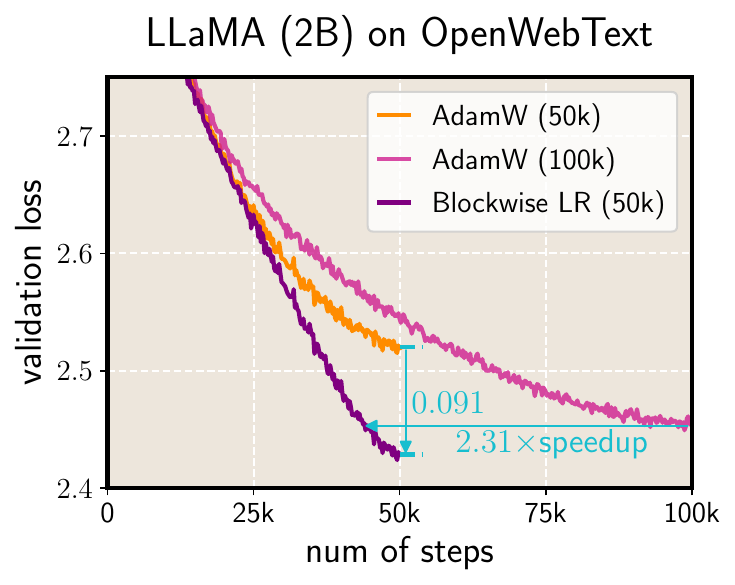}
    \vspace{-.1cm}
    \caption{AdamW with Blockwise LR consistently outperforms AdamW in LLM pre-training tasks across different model types, varying model sizes, and datasets.}
    \label{fig: main results: blockwise lr}
    \vspace{-.0cm}
\end{figure*}

{\bf Main findings.}
In Figure~\ref{fig: main results: blockwise lr} and Figures~\ref{fig: introduction}(right), we compare the performance of AdamW with Blockwise LR against vanilla AdamW across various settings. Our observations, which consistently hold across all experiments--including both GPT-2 and LLaMA models with sizes ranging from 0.12B to 2B--and datasets including OpenWebText and MiniPile, are as follows:

\vspace{-.2cm}

\begin{itemize}
\item Given the same total number of training steps,  Blockwise LR enables AdamW to reach a {\bf lower terminal loss} than vanilla AdamW.
\item Across different total training steps, AdamW with Blockwise LR achieves a  {\bf nearly $2\times$ speedup}  compared to vanilla AdamW.
\vspace{-.2cm}
\end{itemize}

An intriguing observation in Figure~\ref{fig: main results: blockwise lr} is that  AdamW with Blockwise LR often starts to outperform vanilla AdamW  from the mid-to-late stages of training. This behavior resembles the WSD scheduler~\citep{wen2024understanding,hu2024minicpm}, which typically surpasses cosine or linear decay LR schedulers in the late stage (during the decay phase). Understanding the underlying cause of this phenomenon requires further investigation, which we leave for future work.

\begin{figure*}[!ht]
    \centering
    \includegraphics[width=0.27\linewidth]{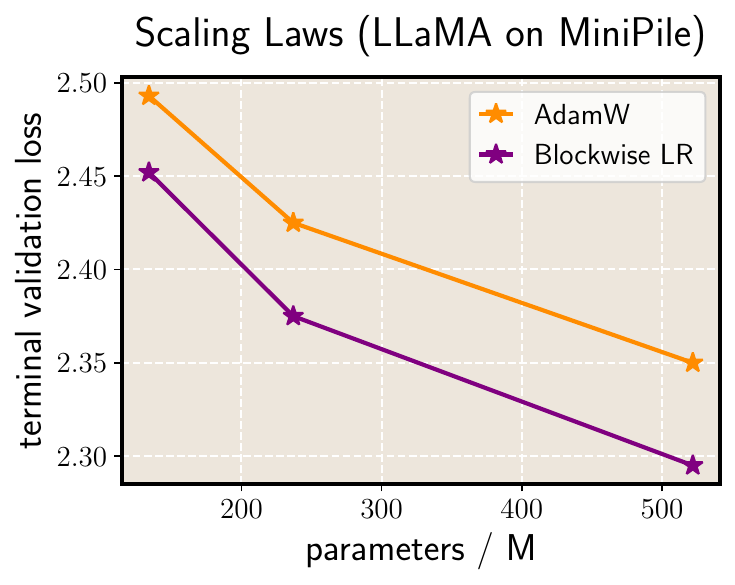}
    \includegraphics[width=0.282\linewidth]{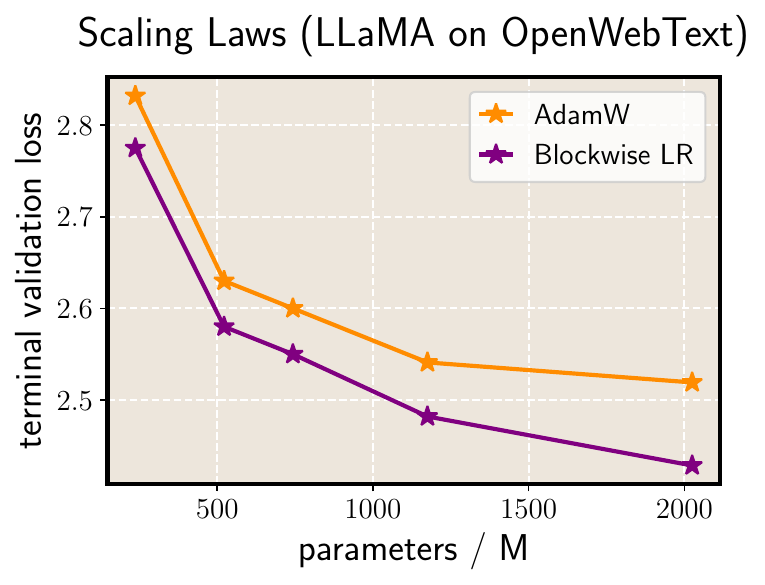}
    \includegraphics[width=0.27\linewidth]{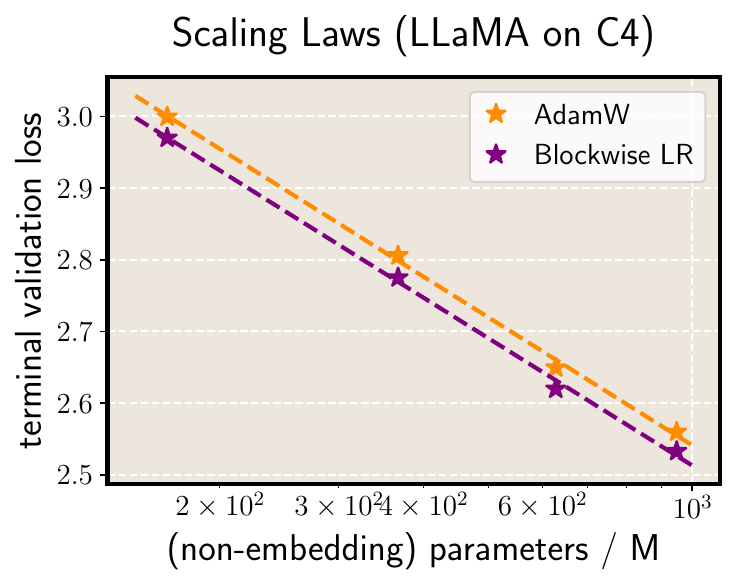}
    \vspace{-.2cm}
    \caption{Scaling-law comparison of AdamW with Blockwise LR and AdamW on various datasets for LLaMA models.}
    \label{fig: scaling law}
    \vspace*{-.0cm}
\end{figure*}

{\bf Scaling law is in favor of Blockwise LR.}
To further examine scaling behavior, Figure~\ref{fig: scaling law} (right) visualizes the scaling laws of AdamW with Blockwise LR versus AdamW during LLaMA pre-training. 
For MiniPile (left) and OpenWebText (middle), the performance gaps between the two optimizers {\em get larger as models size grows}.
For C4 (right), the performance gap remains stable across model scales, with the corresponding scaling curves remaining nearly parallel.
These results suggest that {\em the gains offered by Blockwise LR may persist at larger model scales.}

{\bf Evaluation on downstream tasks.}
Furthermore, as observed in Table~\ref{tab: downstream evaluation},
the improvement in validation loss transfers to an improvement in downstream task accuracy.
Within the same number of pre-training steps, the LLaMA (1.1B) trained with Blockwise LR shows better downstream performance than Adam among all evaluated tasks.

\begin{table*}[tb!]
    \centering
    \caption{Evaluation results on downstream tasks (0-shot with lm-evaluation-harness) of LLaMA models (1.1B) pre-trained on OpenWebText using AdamW or Blockwise LR for 50K steps. The best scores in each column are bolded. 
    }
    \vspace{-.2cm}
    \begin{tabular}{c|c|c|c|c|c|c|c}
    \hline\hline
    Method & \texttt{ARC\_E} & \texttt{ARC\_C} & \texttt{PIQA} & \texttt{HellaSwag} & \texttt{OBQA} & \texttt{WinoGrande} & \texttt{SCIQ} \\\hline
       AdamW  & \small 52.69 &  \small 22.87 &  \small 68.71 &  \small 36.13 &  \small 19.40 &  \small 55.17 &  \small 77.60  \\\hline
       Blockwise LR  &  \small {\bf 54.29} &  \small {\bf 25.34} &  \small {\bf 69.53} &  \small {\bf 38.00} &  \small {\bf 22.60} &  \small {\bf 59.83} &  \small {\bf 81.60} \\\hline\hline
    \end{tabular}
    \label{tab: downstream evaluation}
    \vspace{-.05cm}
\end{table*}


\subsection{Ablation Studies} 


In the preceding experiments, Blockwise LR is applied to all major blocks simultaneously. 
Here, we conduct ablation studies to assess the contribution of each   block type individually. Specifically, we pre-train  a LLaMA model (0.25B) on OpenWebText  focusing on  three comparisons: {\bf (i)} applying Blockwise LR exclusively to \Embed; {\bf (ii)} applying Blockwise LR to both \Embed\ and \FFN; {\bf (iii)} applying Blockwise LR to blocks of all the four types (\Embed, \FFN, \QK, and \VO). 
The adjusting ratios follow Eq.~\eqref{equ: tuned hyperparameters, blockwise lr} and
the results are shown in Table~\ref{tab: ablation studies}. 


First, the results show that applying Blockwise LR to any block consistently improves performance, supporting the hypothesis that dynamics along low-sharpness directions are crucial for loss reduction.
Among all blocks, applying Blockwise LR to \FFN\ yields the largest improvement  ($0.043-0.016=0.027$), likely because \FFN\ blocks comprise the majority of model parameters, offering the greatest potential for optimization gains.


Second, we conduct an additional experiment to assess the impact of increasing the LR for \LN\ blocks. Specifically, the \LN\ LR is doubled, while the LR for other blocks remains unchanged from the baseline. As shown in the last row of Table~\ref{tab: ablation studies}, this leads to a deterioration in performance, contrasting with the improvements seen when increasing the LRs for other blocks by far more than double. This result underscores a fundamental difference in the dynamics of \LN\ with other blocks.

In summary, these ablation studies further validate the effectiveness of Blockwise LR and confirm the rationale of selecting specific types of blocks for LR amplification, as guided by the sharpness disparity principle.

\begin{table}[!htb]
    \centering
    \caption{Ablation results for the effectiveness of Blockwise LR in pre-training LLaMA (0.25B) on OpenWebText.}
    \vspace{-.1cm}
    \begin{tabular}{cl}
   \toprule[1pt]
   \small Blockwise LR & \small terminal loss (50k steps) \\ 
   \midrule[1pt]
   \small  w/o & \small 2.834 \\      \hline
    \small \Embed & \small 2.818 (-0.016 $\text{\ding{51}}$) \\ 
    \small \Embed\ \& \FFN & \small 2.791 (-0.043 $\text{\ding{51}}$) \\
    \small \Embed\ \& \FFN\ \& \QK\ \& \VO & \small 2.784 (-0.050 $\text{\ding{51}}$) \\ \hline
    \small \LN & \small 2.837 (+0.003 $\text{\ding{55}}$) \\ 
    \bottomrule[1pt]
    \end{tabular}
    \label{tab: ablation studies}
    \vspace{-.1cm}
\end{table}

\subsection{Integration into Other Optimization Schemes}

In practice, there are two popular directions for improving LLM pre-training: acceleration and reducing memory consumption. While Blockwise LR has demonstrated remarkable success in accelerating pre-training, a natural {\bf question} arises: {\em Can Blockwise LR be combined with memory-efficient optimizers to achieve both faster training and fewer memory consumption?}

{\bf Blockwise LR on Adam-mini.} Without loss of generality, we choose the Adam-mini~\citep{zhang2024adam} optimizer, an Adam variant  that reduces memory consumption by approximately $2\times$ compared to AdamW. Here, we conduct experiments to explore whether Blockwise LR can also accelerate Adam-mini. Following \citet{zhang2024adam}, we adopt the \texttt{lr\_max} that tuned for AdamW as the the \texttt{lr\_max} of Adam-mini. However, since Adam-mini employs SGD within each block, its dynamics differs significantly from AdamW. Consequently, for Adam-mini with Blockwise LR, we re-tune the ratios $r(\bullet)$ for $\bullet\in\{\Embed,\QK,\FFN,\VO\}$. More experimental details are provided in Appendix~\ref{appendix: experiments for adam-mini}.

\begin{figure}[!h]
    \centering
   \vspace{-.1cm}
    \includegraphics[width=0.495\linewidth]{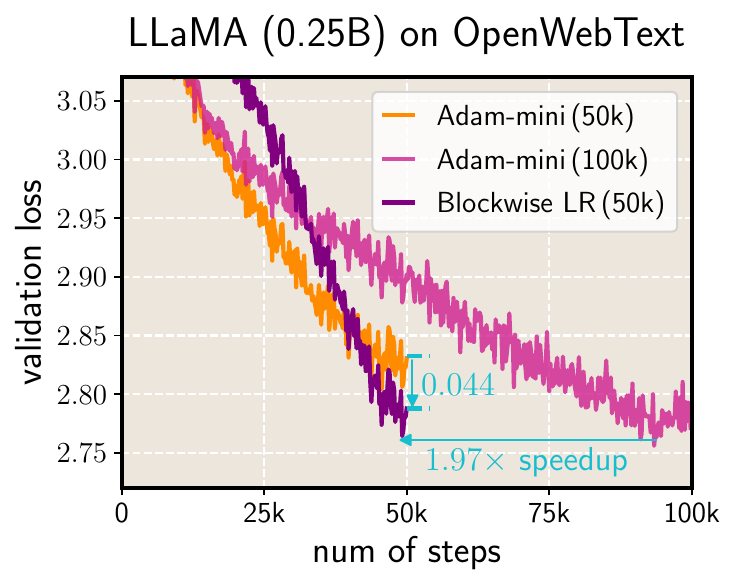}
    \hspace{-.2cm}
    \includegraphics[width=0.495\linewidth]{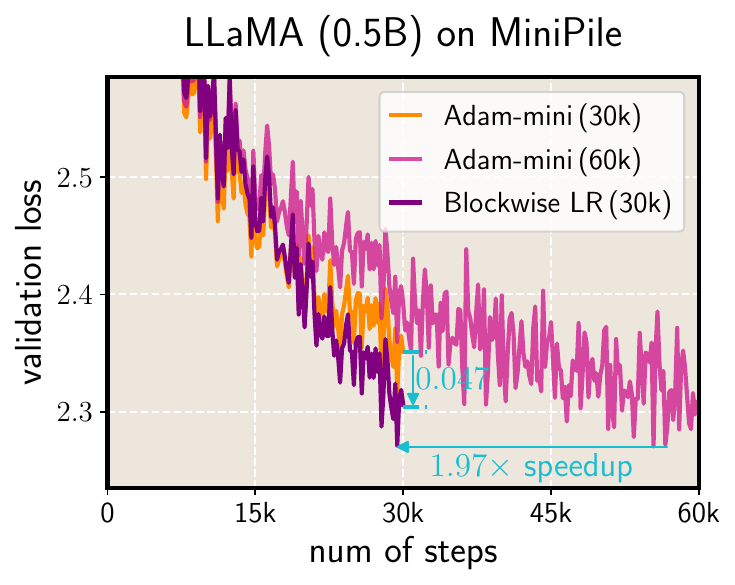}
    \vspace{-.2cm}
    \caption{Adam-mini with Blockwise LR outperforms Adam-mini in pre-training tasks.}
    \label{fig: blockwise lr on adam-mini}
    \vspace{-.1cm}
\end{figure}

The results, presented in Figure~\ref{fig: blockwise lr on adam-mini}, demonstrate that {\bf Blockwise LR  achieves a $2\times$ speedup on Adam-mini}. Since vanilla Adam-mini already achieves a $2\times$ memory saving compared to AdamW while maintaining  nearly the same convergence speed, Adam-mini combined with Blockwise LR achieves both a $2\times$ speedup and $2\times$ memory saving compared to vanilla AdamW. We leave more ablation studies with other optimizers for future work.

\begin{figure}[!ht]
    \centering
    \vspace{-.0cm}
    \includegraphics[width=0.46\linewidth]{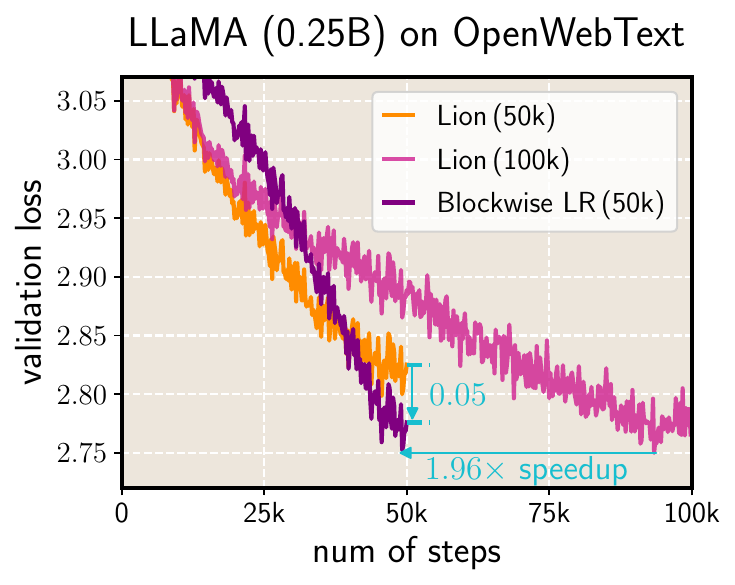}
    \includegraphics[width=0.525\linewidth]{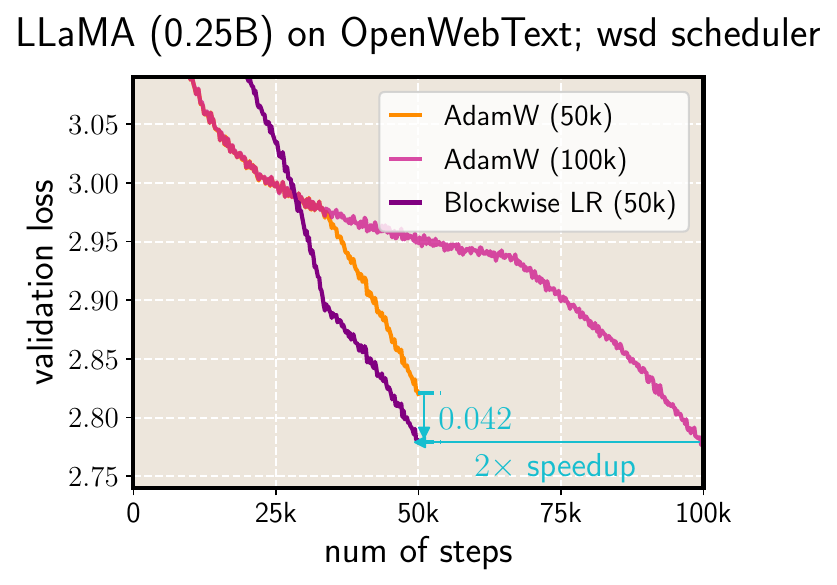}
    \vspace{-.4cm}
    \caption{In pre-training tasks, (left) Lion with Blockwise LR outperforms Lion; (right) when using wsd LR scheduler, AdamW with Blockwise LR outperforms AdamW.}
    \label{fig: blockwise lr on lion or wsd}
    \vspace{-.1cm}
\end{figure}

{\bf Blockwise LR on Lion.} Another memory-efficient optimizer is Lion~\citep{chen2024symbolic}, which eliminates second-order moments in AdamW. We conduct experiments to explore whether Blockwise LR can also accelerate Lion. 
We begin by tuning the \texttt{lr\_max} for Lion baseline, as detailed in Appendix~\ref{appendix: experiments for adam-mini}. 
For Lion with Blockwise LR, we directly apply the ratios $r(\bullet)$ in Eq.~\eqref{equ: tuned hyperparameters, blockwise lr} (note that this is originally tuned for AdamW with Blockwise LR).
The results, presented in Figure~\ref{fig: blockwise lr on lion or wsd} (left), demonstrate that {\em Blockwise LR yields a $2\times$ speedup on Lion}.

Additionally, we evaluate the evolution of the {\em average sharpness across different blocks} when trained using {\em Lion optimizer}.
As shown in Figure~\ref{fig: law llama lion process} in Appendix~\ref{appendix: experiments for adam-mini}, the results {\em closely resemble those} in Figure~\ref{fig: law gpt process}(b), which uses AdamW. Our Principle (Eq.~\eqref{equ: main findings}) emerges during the initial phase, and persists throughout the subsequent training process.

{\bf Blockwise LR with wsd scheduler.}
The preceding experiments employ the cosine decayed LR scheduler. 
In this section, we evaluate Blockwise LR under an alternative and increasingly popular scheduler: warmup-stable-decay (wsd)~\citep{hu2024minicpm}, which includes a linear warm-up LR to peak \texttt{lr\_max}, followed by a stable phase where LR remains at \texttt{lr\_max}, and then a linear decay to \texttt{lr\_min}.
We extend our experiments to incorporate the WSD scheduler. Experimental details are provided in Appendix~\ref{appendix: experiments for adam-mini}. 
As shown in Figure~\ref{fig: blockwise lr on lion or wsd} (right), {\em Blockwise LR still achieves a $2\times$ speedup when used with the wsd scheduler}.

These experiments demonstrate that Blockwise LR is not limited to accelerating AdamW but can also be effectively combined with other optimizers such as Adam-mini and Lion, and LR scheduler such as wsd, while preserving their unique advantages. This finding paves the way for future research exploring the integration of Blockwise LR with other optimization algorithms.

\vspace{-.1cm}

\section{Conclusion and Outlook}


In this paper, we uncovered a sharpness disparity principle among different types of blocks in transformers, as formalized in Eq.~\eqref{equ: main findings}. Notably, this blockwise sharpness disparity persists throughout the entire training process, except during the initial few steps. Building on this discovery, we proposed a novel Blockwise LR adjustment principle, which effectively accelerates base optimizers such as AdamW and Adam-mini in LLM pre-training tasks.  

\textbf{Future works.}
It would be valuable to investigate the applicability of our Blockwise LR to non-LLM tasks, such as computer vision, and its compatibility with other optimizers, such as Muon~\citep{jordan2024muon} and other alloy-like architectures such as Mamba~\citep{gu2023mamba}.
Furthermore, our findings open up opportunities to develop  other block-adaptive optimization strategies, such as blockwise weight decay and gradient clipping, which could further enhance training efficiency and performance.

\section*{Acknowledgements}
Lei Wu was supported by the National Key R$\&$D Program of China (No. 2022YFA1008200) and National Natural Science Foundation of China (No. 12288101). 
Mingze Wang was supported by Young Scientists (PhD) Fund of the National Natural Science Foundation of China (No. 124B2028).
Junchi Yan and Zhanpeng Zhou were partly supported by NSFC (72342023).

\section*{Impact Statement}
This paper contributes to advancing the field of deep learning, with a focus on understanding and improving the pre-training of LLMs. While our work has the potential to impact society in various ways, we do not identify any specific societal consequences that require particular emphasis at this time.

\newpage
\appendix
\onecolumn
\newpage
\appendix



\section{Experimental Details}
\label{sec:experimental_details}

{\bf Models.} We utilize two popular classes of LLM models for our pre-training experiments:
 \begin{itemize}
    \item {\bf GPT-2.} We use GPT-2 (small)  model~\citep{radford2019language}, implemented via the nanoGPT code base ~\citep{Karpathy2022}. 
    Following nanoGPT, the model employs Gaussian Error Linear Unit (GELU) activations and standard Layer Normalization (LayerNorm). Detailed model configurations are provided in Table~\ref{table: model config and max lrs}.
    
    \item {\bf LLaMA.} LLaMA~\citep{touvron2023LLaMA} is another popular decoder-only Transformer architecture, incorporating Rotary Positional Encoding (RoPE)~\citep{su2024roformer}, Swish-Gated Linear Unit (SwiGLU), and Root mean square layer normalization (RMSNorm). 
    We pre-train LLaMA models of sizes ranging from 0.13B to 2B parameters. 
    For implementation, for the 1.1B model configuration, we follow TinyLlama~\citep{zhang2024tinyllama}, which employs grouped-query attention~\citep{ainslie-etal-2023-gqa}; for other model sizes, we utilize the LLaMA code from HuggingFace Transformers Library~\citep{wolf-etal-2020-transformers}. Additional model configurations are detailed in Table~\ref{table: model config and max lrs} and~\ref{table: c4 model config and max lrs}.
\end{itemize}

{\bf Datasets.} Models are pre-trained on the following datasets:

\begin{itemize}
    \item {\bf OpenWebText}~\citep{Gokaslan2019OpenWeb}. It is an opensource recreation of the WebText corpus, is extensively utilized for LLM pre-training such as RoBERTa~\citep{liu2019roberta} and GPT-2. 
    \item {\bf MiniPile.}~\citep{kaddour2023minipile}. It is a 6GB subset of the deduplicated Pile (825GB)~\citep{gao2020pile} presents a highly diverse text corpus. Given its diversity, training on minipile poses challenges and potential instabilities.
    \item{\bf Colossal Clean Crawled Corpus (C4)}~\citep{raffel2020exploring}. It is a large-scale public language dataset, widely used for LLM pre-training such as T5~\citep{raffel2020exploring}, and prior pre-training studies~\citep{zhao2024galore,zhao2024deconstructing}. 
\end{itemize}

All experiments are conducted on 4 A800/H800 80G GPUs.

\subsection{Training Configurations for AdamW Baselines}
\label{appendix: experiments for baselines}

\begin{table}[!ht]
		\centering
        \renewcommand{\arraystretch}{1.25}
		\caption{\small Model configurations and optimally-tuned peak learning rates on OpenWebText and MiniPile.}
		\label{table: model config and max lrs}
		\begin{small}
		\begin{tabular}{l|c|c|c|c|c|c|c}
		\hline 
		Acronym & Size & $d_{\mathrm{model}}$ & $d_{\mathrm{FF}}$ & n$\_$head & depth &  \texttt{lr\_max} on OpenWebText & \texttt{lr\_max} on MiniPile \\
		\hline\hline 
        GPT-2 (small) & 124M & 768 & 3072 & 12 & 12 & 6e-4 & 6e-4 \\
	LLaMA (0.13B) & 134M &  768 & 3072 & 12 & 6 & -- & 1.2e-3 \\
	LLaMA (0.25B) & 237M & 1024 & 4096 & 16 & 8 & 8e-4 & 7.5e-4 \\
	LLaMA (0.5B) & 522M & 1280 & 5120 & 20 & 15 & 8e-4 & 4.5e-4 \\
	LLaMA (0.75B) & 743M & 1664 & 6656 & 26 & 13 & 6e-4 & -- \\
	LLaMA (1.1B) & 1175M & 2048 & 5632 & 32 & 22 & 4e-4 & -- \\ 
    LLaMA (2B) & 2025M & 2048 & 8192 & 32 & 22 & 2e-4 & -- \\ 
	 \hline 
		\end{tabular}
		\end{small}
\end{table}

\begin{table}[!ht]
		\centering
        \renewcommand{\arraystretch}{1.25}
		\caption{\small Model configurations and optimally-tuned peak learning rates on C4.}
		\label{table: c4 model config and max lrs}
		\begin{small}
		\begin{tabular}{l|c|c|c|c|c|c}
		\hline 
		Acronym & Size & $d_{\mathrm{model}}$ & $d_{\mathrm{FF}}$ & n$\_$head & depth &  \texttt{lr\_max} \\\hline\hline 
	LLaMA (66M) & 66M & 512  & 2048 & 8 & 8 & 1e-3 \\
    LLaMA (0.2B) & 200M &  768 & 3072 & 16 & 8 & 1e-3 \\
	LLaMA (0.4B) & 400M & 1280 & 5120 & 16 & 12 & 6e-4 \\
	LLaMA (1B) & 1004M & 1600 & 6400 & 25 & 22 & 3e-4 \\ 
	 \hline 
	\end{tabular}
	\end{small}
\end{table}

As a baseline optimizer, we use the default AdamW for LLM pre-training, configured with the hyperparameters $\beta_1=0.9,\beta_2=0.95$ and weight decay $\lambda=0.1$. To ensure training stability, gradient clipping is applied by norm with threshold $1.0$. These settings align with the training protocols used in nanoGPT and LLaMA models ~\citep{touvron2023LLaMA}.
The default LR strategy integrates a linear warm-up
phase, followed by a cosine decay scheduler with the peak learning rate \texttt{lr\_max} and the final learning rate \texttt{lr\_min}=\texttt{lr\_max}$/20$. Additionally,

\begin{itemize}[leftmargin=2em]

    \item {\bf OpenWebText pre-training.} The (max) sequence length is set to 1024, and the batch size is set to 480, following nanoGPT and~\citet{liu2023sophia}. The total training duration is 50,000 or 100,000 steps, including 1,000 warm-up steps.
    The grid search for \texttt{lr\_max} is performed over $\{$\texttt{2e-4}, \texttt{4e-4}, \texttt{6e-4}, \texttt{8e-4}, \texttt{1e-3}$\}$. Optimal learning rates for each model are detailed in Table~\ref{table: model config and max lrs}.

    \item {\bf MiniPile pre-training.} The (max) sequence length is set to 512, and the batch size is set to 300, following~\citet{wang2024improving}. The total training duration is 30,000 or 60,000 steps, including 600 warm-up steps. The grid search for \texttt{lr\_max} is performed over $\{$\texttt{3e-4}, \texttt{4.5e-4}, \texttt{6e-4}, \texttt{7.5e-4}, \texttt{9e-4}, \texttt{1.2e-3}, \texttt{1.5e-3}$\}$. Optimal learning rates for each model are detailed in Table~\ref{table: model config and max lrs}.

    \item {\bf C4 pre-training} We follow the setup of \citet{zhao2024galore,zhao2024deconstructing}, using a sequence length of 256 and batch size of 512.
    Following the Chinchilla scaling law~\citep{hoffmann2022training}, the total number of training tokens is set to be approximately 20 times the number of model parameters. The training includes 1,000 warm-up steps.
    The grid search for \texttt{lr\_max} is performed over $\{$\texttt{1e-4}, \texttt{2e-4}, \texttt{3e-4}, \texttt{6e-4}, \texttt{1e-3}, \texttt{1.5e-3}$\}$.
    Optimal learning rates for each model are detailed in Tables~\ref{table: c4 model config and max lrs}. We use the T5 tokenizer, with the vocabulary size 32100.
    
\end{itemize}

{\bf Baselines}: models are pre-trained using AdamW with the respective tuned \texttt{lr\_max} for each dataset and model configuration.

\begin{figure}[!ht]
    \centering
    \subfloat[
    (To illustrate Theorem~\ref{thm: FFN vs Norm}) Norms of input/output weight parameters in $\FFN$ and the weight parameters of $\Norm$ before $\FFN$, averaged by the number of layers.]{\includegraphics[width=0.25\linewidth]{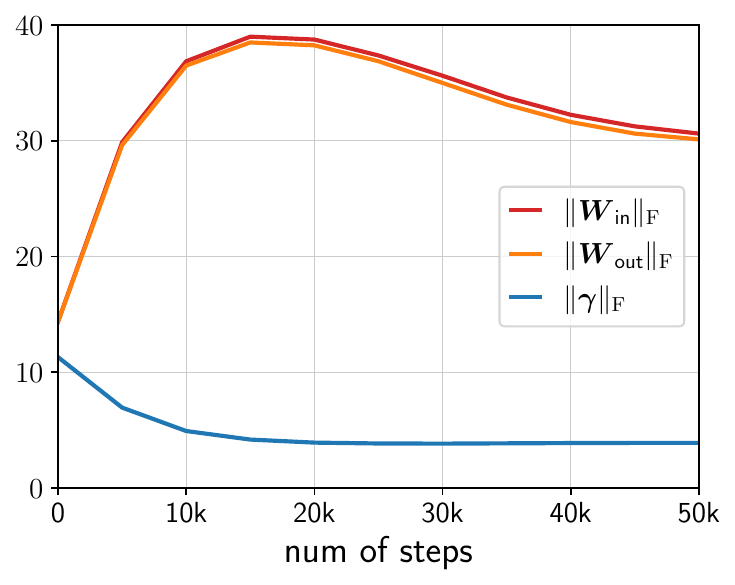}}
    \hspace{.4cm}
    \subfloat[To illustrate Theorem~\ref{thm: QK VO vs Norm}) Norms of query/key/value/output parameters in $\SA$ and the weight parameters of \Norm\ before $\SA$, averaged by the number of layers.]{\includegraphics[width=0.25\linewidth]{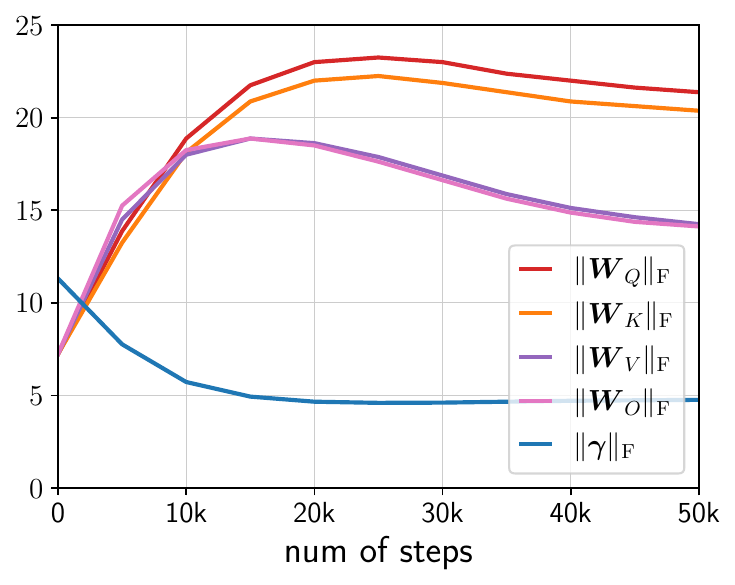}}
    \hspace{.4cm}
    \subfloat[(To illustrate Theorem~\ref{thm: Embed vs Norm}) Norms of weight parameters in $\Embed$ and the weight parameters in the adjoint \Norm\ layer after $\Embed$.]
    {\includegraphics[width=0.25\linewidth]{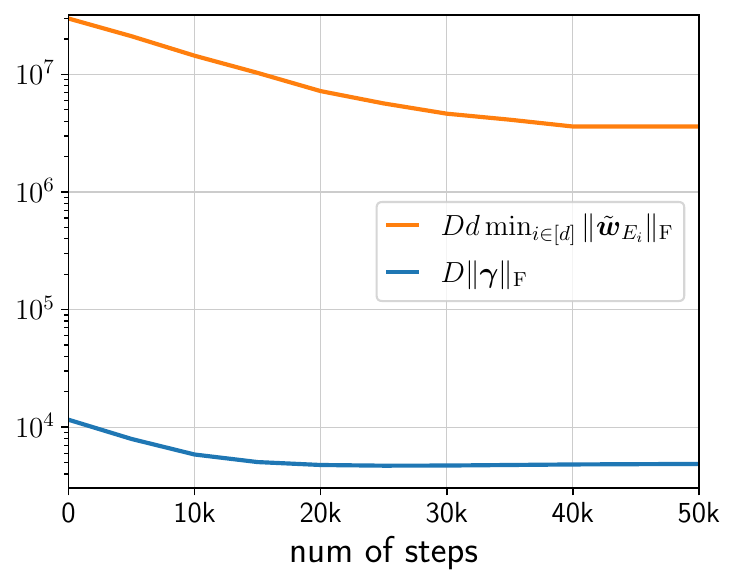}}
    \caption{Evolution of parameter norms across different blocks during pre-training LLaMA (0.25B) on OpenWebText.}
    \label{fig: blockwise norm}
\end{figure}

\begin{figure}[!ht]
    \centering
    \subfloat[Average sharpness across different layers. Layer $0$ corresponds to the \Embed\ layer. Layers $1,\cdots,8$ correspond to the \SA-\FFN\ layers.]{\includegraphics[width=0.3\linewidth]{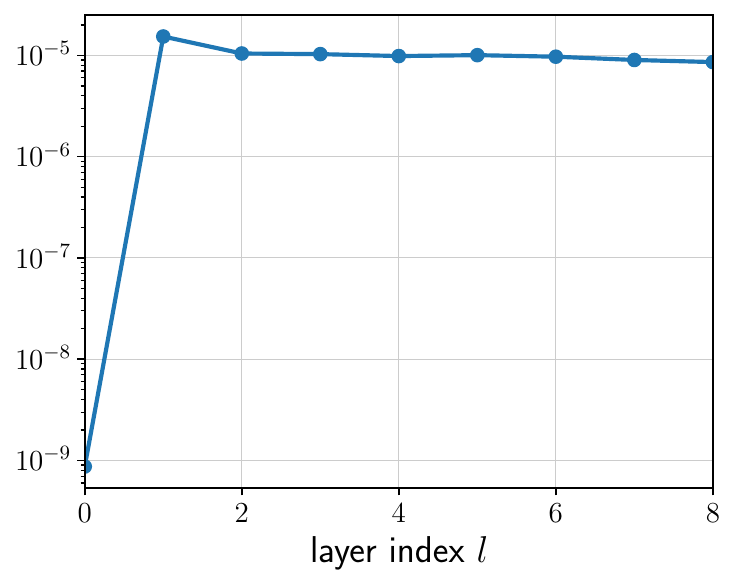}}
    \hspace{.4cm}
    \subfloat[Average sharpness of the blocks ($\bullet\in\{\QK,\FFN,\VO,\Norm\}$) across different layers ($l=1,\cdots,8$).]
    {\includegraphics[width=0.3\linewidth]{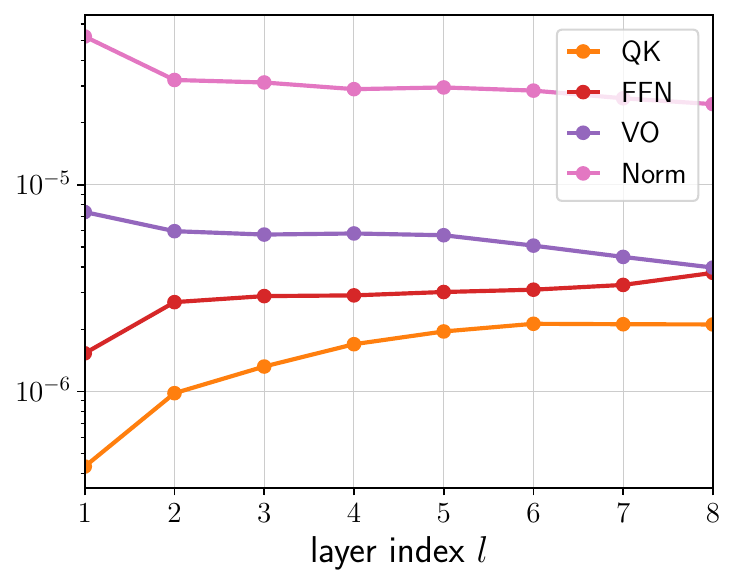}}
    \caption{In a pre-trained LLaMA (0.25B) (with $L=8$ layer), there is no clear disparity for the average sharpness across the {\bf layers}. This is in stark contrast to our our sharpness disparity {\bf Principle}~\eqref{equ: main findings} across the {\bf blocks}.}
    \label{fig: layerwise no law}
\end{figure}

{\bf Related Experiments.}
\begin{itemize}[leftmargin=2em]
    \item {\bf Blockwise LR Experiments.} The baseline results in {\bf Figure~\ref{fig: main results: blockwise lr}}, {\bf Figure~\ref{fig: introduction} (right)}, and {\bf Table~\ref{tab: ablation studies} (the w/o line)} are trained following the configurations above.
    
    \item {\bf Sharpness Principle Experiments.} Models for {\bf Figure~\ref{fig: introduction} (left)}, {\bf Figure~\ref{fig: law final}}, {\bf Figure~\ref{fig: law gpt process}}, are trained using the baseline configurations for GPT-2 (small) or LLaMA (0.25B) on OpenWebText, with a total training duration 50,000 steps.
    In these experiments, the sharpness is estimated using $\bh(\btheta)$ in Eq.~\eqref{equ: fisher estimate}, with $B$ set to 1024. The sharpness distributions and average sharpness values for different blocks ($\bullet$) are calculated on a logarithmic scale, i.e., $\log \bh(\btheta[\bullet])$.

    Additionally, the experiment in Figure~\ref{fig: layerwise no law} employs the same model and sharpness estimator.
    
    \item {\bf Theoretical Analysis Support.} To support our theoretical insights in Section~\ref{section: theoretical insights}, {\bf Figure~\ref{fig: blockwise norm}} shows the evaluation of the parameter norms across different blocks during training. The model used is LlaMa (0.25B), trained on OpenWebText. The model is LLaMA (0.25B), trained on OpenWebText following the baseline configurations.
\end{itemize}

\subsection{Experimental Details for Blockwise LR on AdamW}
\label{appendix: experiments, blockwise LR}

{\bf Switching Time.} The principle of blockwise sharpness heterogeneity emerges clearly after the initial training phase, as shown in Figure~\ref{fig: law gpt process}. To leverage this principle, in our experiments of AdamW using Blockwise LR, we {\bf switch} from standard AdamW to AdamW with Blockwise LR {\bf at the end of LR warmup phase}. 

{\bf Experiments in Figure~\ref{fig: main results: blockwise lr}.} We adopt the adjusting ratios~\eqref{equ: tuned hyperparameters, blockwise lr} as the default adjusting ratios for {\bf} all experiments of AdamW with Blockwise LR.

\begin{figure}[!ht]
    \centering
    \includegraphics[width=0.35\linewidth]{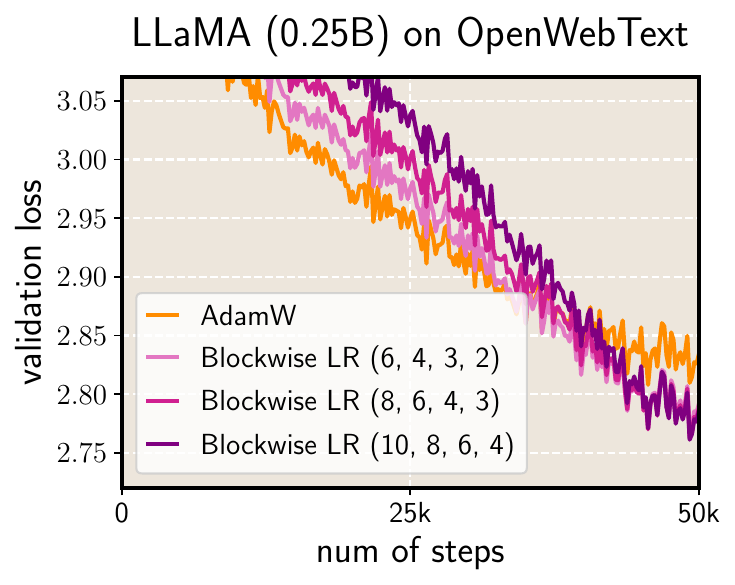}
    \caption{Pre-training LLaMA (0.25B) on Minipile using AdamW with Blockwise LR across three configurations of adjusting ratios.}
    \label{fig: robust llama web}
\end{figure}

{\bf Experiment on Hyper-parameter Tuning.} We {\bf only} tune the four adjusting ratios $r(\bullet)$ ($\bullet\in\{\Embed,\QK,\VO,\FFN\}$) in a single small-scale experiment: pre-training LLaMA (0.25B) on Minipile. Specifically, we compare the results under the following configurations of ratios:
\begin{gather*}
r(\Embed)=6, r(\QK)=4, r(\FFN)=3, r(\VO)=2;\\
r(\Embed)=8, r(\QK)=6, r(\FFN)=4, r(\VO)=3;\\
r(\Embed)=10, r(\QK)=8, r(\FFN)=6, r(\VO)=4.
\end{gather*}
The results for the tuning experiments are presented in {\bf Figure~\ref{fig: robust llama web}}. One can see that the configuration $r(\Embed)=10, r(\QK)=8, r(\FFN)=6, r(\VO)=4$ (Eq.~\eqref{equ: tuned hyperparameters, blockwise lr}) achieves the largest improvement in terminal loss.
Additionally, Blockwise LR demonstrates robustness to these ratios, consistently accelerating pre-training across all tested configurations.

{\bf Experiments in Table~\ref{tab: ablation studies}.} We pre-train LLaMA (0.25B) on
OpenWebText with a focusing on the three comparisons: (i) applying Blockwise LR exclusively to \Embed; (ii) applying Blockwise LR to both \Embed\ and \FFN; (iii) applying Blockwise LR to blocks of all the four types (\Embed, \FFN, \QK, and \VO). The adjusting ratios are maintained as per the tuned in Eq.~\eqref{equ: tuned hyperparameters, blockwise lr}.

\subsection{Experimental details for Adam-mini, Lion, and wsd}
\label{appendix: experiments for adam-mini}

{\bf Experiments for Adam-mini.}

\begin{itemize}[leftmargin=2em]
    \item {\bf Baseline.} In the baseline experiments in Figure~\ref{fig: blockwise lr on adam-mini}, following~\citet{zhang2024adam}, we adopt the same peak learning rate \texttt{lr\_max} tuned for AdamW as the \texttt{lr\_max} of Adam-mini.
    
    \item {\bf Hyperparameter tuning.} Since Adam-mini uses SGD within each blocks, its dynamics {\bf differs significantly} from those of AdamW. Thus, for Adam-mini with Blockwise LR, we re-tune the ratios $r(\bullet)\in\{1,2,4\}$ for $\bullet\in\{\Embed,\QK,\FFN,\VO\}$. The tuned ratios are $r(\Embed)=4, r(\QK)=1, r(\FFN)=4, r(\VO)=4$, which are used in the experiments in {\bf Figure~\ref{fig: blockwise lr on adam-mini}}.
Note that these ratios do not satisfy $r(\bullet)\propto\frac{\cS(\Norm)}{\cS(\bullet)}$. This discrepancy may stem from the unique dynamics of Adam-mini, particularly its SGD-like behavior within blocks. We leave further investigation for future work.

\end{itemize}

{\bf Experiments for Lion}
In the baseline experiments in Figure~\ref{fig: blockwise lr on lion or wsd} (left), following~\citet{chen2024symbolic}, we search for the optimal maximum learning rate \texttt{lr\_max} for Lion within the set $\{1/3, 1/5, 1/10\}$  of the \texttt{lr\_max} values tuned for AdamW, using corresponding weight decay values of $\{$0.3, 0.5, 1.0$\}$. Ultimately, we adopt 1/5 of the AdamW-tuned \texttt{lr\_max} with a weight decay of 0.5.
For Lion with Blockwise LR, we directly apply the ratios in Eq.~\eqref{equ: tuned hyperparameters, blockwise lr} (note that this is originally tuned for AdamW with Blockwise LR).

\begin{figure}[ht!]
    \centering
    \includegraphics[width=0.95\linewidth]{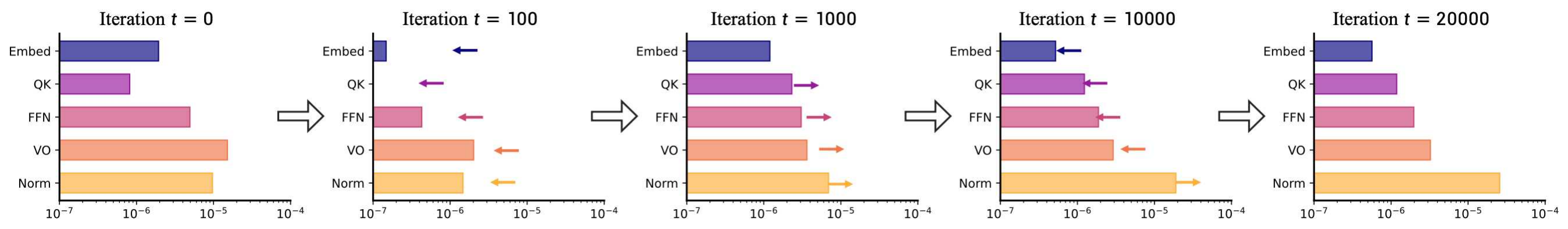}
    \vspace{-.15cm}
    \caption{Evolution of the average sharpness across different blocks during pre-training LLaMA (0.25B) on OpenWebText using the {\bf Lion optimizer}. The total training duration is 50k steps. 
    The results closely {\em resemble those} in our Figure~\ref{fig: law gpt process}(b), which uses AdamW.
    Our Principle (Eq.~\eqref{equ: main findings}) emerges during the initial phase (from iteration 0 to iteration 1k), which accounts for only approximately $2\%$ of the total steps, and persists throughout the subsequent training process.}
    \label{fig: law llama lion process}
\end{figure}

{\bf Experiments for wsd sheduler.} In the experiments in Figure~\ref{fig: blockwise lr on lion or wsd} (right), we use a linear warm-up LR to peak \texttt{lr\_max}, followed by a stable phase where LR remains at \texttt{lr\_max} (up to 66.7\% of the total training steps), and then a linear decay to 0. The \texttt{lr\_max} is identical to that used in the cosine decay scheduler, as reported in Table~\ref{table: model config and max lrs}.

\vspace{1.cm}

\section{Proofs in Section~\ref{section: principle}}
\label{appendix: proof: section principle}

\subsection{Proof of Theorem~\ref{thm: FFN vs Norm}}
\label{appendix: proof: thm: FFN vs Norm}

We focus on the transformation from $\bX^{(l-1)}$ to $\bX^{(l-1/2)}$:
$$\bX^{(l)}=\bX^{(l-1/2)}+\FFN^{(l)}\left(\LN^{l}\left(\bX^{(l-1/2)};\bgamma^{(l)}\right);\bW_1^{(l)},\bW_2^{(l)}\right).$$
From the chain rule, it follows that:
\begin{align*}
    \frac{\partial \cQ}{\partial \bW_{\bullet}^{(l)}}&=\frac{\partial \cQ}{\partial \bX^{(l)}}\frac{\partial \bX^{(l)}}{\partial \bW_{\bullet}^{(l)}},\quad \bullet\in\{1,2\};
    \\
    \frac{\partial \cQ}{\partial \bgamma^{(l)}}&=\frac{\partial \cQ}{\partial \bX^{(l)}}\frac{\partial \bX^{(l)}}{\partial\bgamma^{(l)}}.
\end{align*}

Thus, it suffices to compute $\frac{\partial \bX^{(l)}}{\partial\bW_{\bullet}^{(l)}}$ and $\frac{\partial \bX^{(l)}}{\partial\bgamma^{(l)}}$.
For simplicity, we define:
\begin{gather*}
    \bX:=\bX^{(l-1/2)},\quad
    \bX_{\rm std}=\frac{\bX-\bbE_r[\bX]}{\sqrt{\bbV_r[\bX]}},\quad
    \bX_\LN:=\LN(\bX;\bgamma)=\bX_{\rm std}\odot(\bone_{n\times 1}\otimes\bgamma),
    \\
    \MSA:=\bX_{\LN}\bW_1,\quad
    \ASA:=\sigma(\MSA),\quad
    \FFFN:=\ASA\bW_2,
    \quad
    \bY:=\bX^{(l)}=\bX+\FFFN,
\end{gather*}

where $\sigma(\cdot)$ represents the ReLU or Leacky ReLU activation function.
We now compute $\frac{\partial \bY}{\partial\bW_{\bullet}}$ and $\frac{\partial \bY}{\partial\bgamma}$.

It is straightforward that:
\begin{align*}
    \frac{\partial \bY}{\partial\bW_{1}}=&\frac{\partial\FFFN}{\partial\bW_{1}}=\frac{\partial\FFFN}{\partial\ASA}\frac{\partial\ASA}{\partial\MSA}\frac{\partial\MSA}{\partial\bW_{1}}=\left(\bI_{n}\otimes\bW_2^\top\right)\frac{\partial\ASA}{\partial\MSA}\left(\bX_{\LN}\otimes\bI_M\right);
\end{align*}
\begin{align*}
    \frac{\partial \bY}{\partial\bgamma}=&\frac{\partial\FFFN}{\partial\bX_{\LN}}\frac{\partial\bX_{\LN}}{\partial\bgamma}=\frac{\partial\FFFN}{\partial\ASA}\frac{\partial\ASA}{\partial\MSA}\frac{\partial\MSA}{\partial\bX_{\LN}}\frac{\partial\bX_{\LN}}{\partial\bgamma}
    \\=&\left(\bI_{n}\otimes\bW_2^\top\right)\frac{\partial\ASA}{\partial\MSA}\left(\bI_n\otimes\bW_1^\top\right)\Big(\diag\big(\vec(\bX_{\rm std})\big)\big(\bone_{n\times 1}\otimes \bI_{D}\big)\Big).
\end{align*}

For the (Leaky) ReLU, it holds that $\sigma(z)=z\sigma'(z)$. Thus, for $\frac{\partial \bY}{\partial\bW_2}$, we have: 
\begin{align*}
    \frac{\partial \bY}{\partial\bW_{2}}=&\frac{\partial\FFFN}{\partial\bW_{2}}=\ASA\otimes\bI_D=\left(\bX_{\LN}\bW_1\odot\frac{\partial\ASA}{\partial\MSA}\right)\otimes\bI_D.
\end{align*}

Now we derive the upper bounds. First, notice that:
\begin{align*}
    \norm{\bX_{\rm std}}_\rF=\left(\sum_{i=1}^n \left(\frac{\bX_{i,:}-\bbE[\bX_{i,:}]}{\sqrt{\bbV[\bX_{i,:}]}}\right)^2\right)^{1/2}=\left(\sum_{i=1}^n D\right)^{1/2}=\sqrt{nD};
\end{align*}
\begin{align*}
    &\norm{\bX_\LN}_\rF
    =\norm{\bX_{\rm std}\odot(\bone_{n\times 1}\otimes\bgamma)}_\rF
    \leq\norm{\bX_{\rm std}}_\rF\norm{\bone_{n\times 1}\otimes\bgamma}_\rF\leq\sqrt{nD}\norm{\bone_{n\times 1}}_\rF\norm{\bgamma}_\rF\leq n\sqrt{D}\norm{\bgamma}_\rF.
\end{align*}

Consequently, we have the following estimates:
\begin{align*}
    &\norm{\frac{\partial \cQ}{\partial\bW_{1}}}_\rF
    \leq\norm{\frac{\partial \cQ}{\partial\bY}}_\rF\norm{\frac{\partial \bY}{\partial\bW_{1}}}_\rF
    =\norm{\frac{\partial \cQ}{\partial\bY}}_\rF\norm{\left(\bI_{n}\otimes\bW_2^\top\right)\frac{\partial\ASA}{\partial\MSA}\left(\bX_{\LN}\otimes\bI_M\right)}_\rF\\
    \leq&\norm{\frac{\partial \cQ}{\partial\bY}}_\rF\norm{\frac{\partial\ASA}{\partial\MSA}}_\rF\norm{\bI_{n}\otimes\bW_2^\top}_2\norm{\bX_{\LN}\otimes\bI_M}_2
    \leq\norm{\frac{\partial \cQ}{\partial\bY}}_\rF\norm{\frac{\partial\ASA}{\partial\MSA}}_\rF\norm{\bI_{n}}_2\norm{\bI_M}_2\norm{\bW_2^\top}_\rF\norm{\bX_{\LN}}_\rF
    \\\leq&\norm{\frac{\partial \cQ}{\partial\bY}}_\rF\norm{\frac{\partial\ASA}{\partial\MSA}}_\rF\norm{\bW_2}_\rF\norm{\bX_{\LN}}_\rF\leq n\sqrt{D}\norm{\frac{\partial \cQ}{\partial\bY}}_\rF\norm{\frac{\partial\ASA}{\partial\MSA}}_\rF\norm{\bW_2}_\rF\norm{\bgamma}_\rF;
\end{align*}
\begin{align*}
    &\norm{\frac{\partial \cQ}{\partial\bW_{2}}}_\rF
    \leq\norm{\frac{\partial \cQ}{\partial\bY}}_\rF\norm{\frac{\partial \bY}{\partial\bW_{2}}}_\rF
    =\norm{\frac{\partial \cQ}{\partial\bY}}_\rF\norm{\left(\bX_{\LN}\bW_1\odot\frac{\partial\ASA}{\partial\MSA}\right)\otimes\bI_D}_\rF\\
    \leq&\norm{\frac{\partial \cQ}{\partial\bY}}_\rF\norm{\left(\bX_{\LN}\bW_1\odot\frac{\partial\ASA}{\partial\MSA}\right)}_\rF\norm{\bI_D}_2
    \leq\norm{\frac{\partial \cQ}{\partial\bY}}_\rF\norm{\frac{\partial\ASA}{\partial\MSA}}_\rF\norm{\bX_{\LN}\bW_1}_\rF
    \\\leq&\norm{\frac{\partial \cQ}{\partial\bY}}_\rF\norm{\frac{\partial\ASA}{\partial\MSA}}_\rF\norm{\bW_1}_\rF\norm{\bX_{\LN}}_\rF\leq n\sqrt{D}\norm{\frac{\partial \cQ}{\partial\bY}}_\rF\norm{\frac{\partial\ASA}{\partial\MSA}}_\rF\norm{\bW_1}_\rF\norm{\bgamma}_\rF;
\end{align*}
\begin{align*}
    &\norm{\frac{\partial \cQ}{\partial\bgamma}}_\rF
    \leq\norm{\frac{\partial \cQ}{\partial\bY}}_\rF\norm{\frac{\partial \bY}{\partial\bgamma}}_\rF
    \\=&\norm{\frac{\partial \cQ}{\partial\bY}}_\rF\norm{\left(\bI_{n}\otimes\bW_2^\top\right)\frac{\partial\ASA}{\partial\MSA}\left(\bI_n\otimes\bW_1^\top\right)\Big(\diag\big(\vec(\bX_{\rm std})\big)\big(\bone_{n\times 1}\otimes \bI_{D}\big)\Big)}_\rF
    \\\leq&\norm{\frac{\partial \cQ}{\partial\bY}}_\rF\norm{\frac{\partial\ASA}{\partial\MSA}}_\rF\norm{\bI_{n}\otimes\bW_2^\top}_\rF\norm{\bI_{n}\otimes\bW_1^\top}_\rF\norm{\diag\big(\vec(\bX_{\rm std})\big)\big(\bone_{n\times 1}\otimes \bI_{D}\big)}_\rF
    \\\leq&
    \norm{\frac{\partial \cQ}{\partial\bY}}_\rF\norm{\frac{\partial\ASA}{\partial\MSA}}_\rF\norm{\bI_{n}}_2\norm{\bW_2}_\rF\norm{\bI_{n}}_2\norm{\bW_1}_\rF\norm{\diag\big(\vec(\bX_{\rm std})\big)}_\rF\norm{\bone_{n\times 1}\otimes \bI_{D}}_\rF
    \\\leq&
    \norm{\frac{\partial \cQ}{\partial\bY}}_\rF\norm{\frac{\partial\ASA}{\partial\MSA}}_\rF\norm{\bW_1}_\rF\norm{\bW_2}_\rF\norm{\bX_{\rm std}}_\rF\norm{\bone_{n\times 1}}_\rF\norm{\bI_{D}}_2
    \\\leq& n\sqrt{D}\norm{\frac{\partial \cQ}{\partial\bY}}_\rF\norm{\frac{\partial\ASA}{\partial\MSA}}_\rF\norm{\bW_1}_\rF\norm{\bW_2}_\rF.
\end{align*}

Thus, if we define
$$\Psi:=n\sqrt{D}\norm{\frac{\partial \cQ}{\partial\bY}}_\rF\norm{\frac{\partial\ASA}{\partial\MSA}}_\rF\norm{\bW_1}_\rF\norm{\bW_2}_\rF\norm{\bgamma}_\rF,$$

then it holds that:
\begin{gather*}
\norm{\frac{\partial \cQ}{\partial\bW_1}}_\rF\leq\frac{\Psi}{\norm{\bW_1}_{\rF}};\quad
\norm{\frac{\partial \cQ}{\partial\bW_2}}_\rF\leq\frac{\Psi}{\norm{\bW_2}_{\rF}};\quad
\norm{\frac{\partial \cQ}{\partial\bgamma}}_\rF\leq\frac{\Psi}{\norm{\bgamma}_{\rF}}
\end{gather*}

Therefore,
\begin{gather*}
    \cS(\bW_{\bullet})=\frac{1}{\#(\bW_{\bullet})}\norm{\frac{\partial \cQ}{\partial\bW_{\bullet}}}_\rF^2=\cO\left(\frac{\Psi^2}{D^2\norm{\bW_\bullet}_\rF^2}\right),\quad\bullet\in\{1,2\};
    \\
    \cS(\bgamma)=\frac{1}{\#(\bgamma)}\norm{\frac{\partial \cQ}{\partial\bgamma}}_\rF^2=\cO\left(\frac{\Psi^2}{D\norm{\bgamma}_\rF^2}\right).
\end{gather*}

\subsection{Proof of Theorem~\ref{thm: QK VO vs Norm}}
\label{appendix: proof: thm: QK VO vs Norm}

We focus on the transformation from $\bX^{(l-1)}$ to $\bX^{(l-1/2)}$: 
$$\bX^{(l-1/2)}=\bX^{(l-1)}+\SA^{(l)}\Big(\LN^{(l-1/2)}\left(\bX^{(l-1)};\bgamma^{(l-1/2)}\right);\bW_K^{(l)},\bW_Q^{(l)},\bW_V^{(l)},\bW_O^{(l)}\Big).$$

From the chain rule, it follows that:
\begin{align*}
    \frac{\partial \cQ}{\partial \bW_{\bullet}^{(l)}}&=\frac{\partial \cQ}{\partial \bX^{(l-1/2)}}\frac{\partial \bX^{(l-1/2)}}{\partial \bW_{\bullet}^{(l)}},\quad \bullet\in\{K,Q,V,O\};
    \\
    \frac{\partial \cQ}{\partial \bgamma^{(l-1/2)}}&=\frac{\partial \cQ}{\partial \bX^{(l-1/2)}}\frac{\partial \bX^{(l-1/2)}}{\partial\bgamma^{(l-1/2)}}.
\end{align*}

Thus, it suffices to compute $\frac{\partial \bX^{(l-1/2)}}{\partial\bW_{\bullet}^{(l-1/2)}}$ and $\frac{\partial \bX^{(l-1/2)}}{\partial\bgamma^{(l-1/2)}}$. For simplicity, we define:
\begin{gather*}
    \bX:=\bX^{(l-1)},\quad
    \bX_{\rm std}=\frac{\bX-\bbE_r[\bX]}{\sqrt{\bbV_r[\bX]}},\quad
    \bX_\LN:=\LN(\bX;\bgamma)=\bX_{\rm std}\odot\bgamma,
    \\
    \MSA:=\frac{\bX_{\LN}\bW_Q\bW_K^\top\bX_{\LN}^\top}{\sqrt{D}},\quad
    \ASA:=\sm\left(\MSA\right),\quad
    \SSA:=\ASA\bX_{\LN}\bW_V\bW_O,
    \\
    \bY:=\bX^{(l-1/2)}=\bX+\SSA.
\end{gather*}
 
We now compute $\frac{\partial \bY}{\partial\bW_{\bullet}}$ and $\frac{\partial \bY}{\partial\bgamma}$:
\begin{align*}
    \frac{\partial \bY}{\partial\bW_{Q}}=&\frac{\partial\SSA}{\partial\bW_{Q}}=\frac{\partial\SSA}{\partial\ASA}\frac{\partial\ASA}{\partial\MSA}\frac{\partial\MSA}{\partial\bW_{Q}}
    =\bracket{\bI_n\otimes \bW_O^\top\bW_V^\top\bX_{\LN}^\top}\frac{\partial\ASA}{\partial\MSA}\bracket{\frac{\bX_{\LN}\otimes \bX_{\LN} \bW_K}{\sqrt{D}}};
\end{align*}
\begin{align*}
    \frac{\partial \bY}{\partial\bW_{K}}=\frac{\partial\SSA}{\partial\bW_{K}}=\frac{\partial\SSA}{\partial\ASA}\frac{\partial\ASA}{\partial\MSA}\frac{\partial\MSA}{\partial\bW_{K}}
    =\bracket{\bI_n\otimes \bW_O^\top\bW_V^\top\bX_{\LN}^\top}\frac{\partial\ASA}{\partial\MSA}\bracket{\frac{\bX_{\LN}\otimes \bX_{\LN} \bW_Q}{\sqrt{D}}};
\end{align*}
\begin{align*}
    \frac{\partial \bY}{\partial\bW_{V}}=\frac{\partial\SSA}{\partial\bW_{V}}=\ASA\bX_{\LN}\otimes\bW_O^\top;
\end{align*}
\begin{align*}
    \frac{\partial \bY}{\partial\bW_{O}}=\frac{\partial\SSA}{\partial\bW_{O}}=\ASA\bX_{\LN}\bW_V\otimes\bI_D.
\end{align*}

Moreover,
\begin{align*}
    &\frac{\partial \bY}{\partial\bgamma}=\frac{\partial \bY}{\partial\bX_{\LN}}\frac{\partial \bX_{\LN}}{\partial\bgamma}
    =\frac{\partial\SSA}{\partial\bX_{\LN}}\frac{\partial \bX_{\LN}}{\partial\bgamma}
    \\=&\Bigg(\frac{1}{\sqrt{D}}\Big(\bI_n\otimes\bW_O^\top\bW_V^\top\bX_{\LN}^\top\Big)\frac{\partial\ASA}{\partial\MSA}\left(\Big(\bI_n\otimes\bX_{\LN}\bW_K\bW_Q^\top\right)+\bK_{n,n} \left(\bI_n\otimes\bX_{\LN}\bW_Q\bW_K^\top\right)\Big)
    \\&\quad+\ASA\otimes\bW_O^\top\bW_V^\top\Bigg)\Big(\diag\big(\vec(\bX_{\rm std})\big)\big(\bone_{n\times 1}\otimes \bI_{d}\big)\Big),
\end{align*}
where $\bK_{n,n}$ is the commutation matrix\footnote{The commutation matrix $\bK_{m,n}$ transforms column-wise vectorization into row-wise vectorization.}.

Recalling the proof in Appendix~\ref{appendix: proof: thm: FFN vs Norm}, we have:
\begin{align*}
    \norm{\bX_{\rm std}}_\rF=\sqrt{nD},
    \quad\norm{\bX_\LN}_\rF\leq n\sqrt{D}\norm{\bgamma}_\rF.
\end{align*}

Then, similar to the proof in Appendix~\ref{appendix: proof: thm: FFN vs Norm}, we have the following upper bounds:
\begin{align*}
    \norm{\frac{\partial \cQ}{\partial\bW_{Q}}}_\rF\leq&
    \frac{1}{\sqrt{D}}\norm{\frac{\partial \cQ}{\partial\bY}}_\rF\norm{\frac{\partial\ASA}{\partial\MSA}}_\rF \norm{\bW_K}_\rF\norm{\bW_V}_\rF\norm{\bW_O}_\rF\norm{\bX_{\LN}}_\rF^3
    \\\leq&
    \frac{(n\sqrt{D})^3}{\sqrt{D}}\norm{\frac{\partial \cQ}{\partial\bY}}_\rF\norm{\frac{\partial\ASA}{\partial\MSA}}_\rF \norm{\bW_K}_\rF\norm{\bW_V}_\rF\norm{\bW_O}_\rF\norm{\bgamma}_\rF^3;
\end{align*}
\begin{align*}
    \norm{\frac{\partial \cQ}{\partial\bW_{K}}}_\rF\leq&
    \frac{1}{\sqrt{D}}\norm{\frac{\partial \cQ}{\partial\bY}}_\rF\norm{\frac{\partial\ASA}{\partial\MSA}}_\rF \norm{\bW_Q}_\rF\norm{\bW_V}_\rF\norm{\bW_O}_\rF\norm{\bX_{\LN}}_\rF^3
    \\\leq&
    \frac{(n\sqrt{D})^3}{\sqrt{D}}\norm{\frac{\partial \cQ}{\partial\bY}}_\rF\norm{\frac{\partial\ASA}{\partial\MSA}}_\rF \norm{\bW_Q}_\rF\norm{\bW_V}_\rF\norm{\bW_O}_\rF\norm{\bgamma}_\rF^3;
\end{align*}
\begin{align*}
    \norm{\frac{\partial \cQ}{\partial\bW_{V}}}_\rF\leq\norm{\frac{\partial \cQ}{\partial\bY}}_\rF
    \norm{\ASA}_\rF\norm{\bW_O}_\rF\norm{\bX_{\LN}}_\rF
    \leq n\sqrt{D}\norm{\frac{\partial \cQ}{\partial\bY}}_\rF
    \norm{\ASA}_\rF\norm{\bW_O}_\rF\norm{\bgamma}_\rF;
\end{align*}
\begin{align*}
    \norm{\frac{\partial \cQ}{\partial\bW_{O}}}_\rF\leq
    \norm{\frac{\partial \cQ}{\partial\bY}}_\rF
    \norm{\ASA}_\rF\norm{\bW_V}_\rF\norm{\bX_{\LN}}_\rF
    \leq n\sqrt{D}\norm{\frac{\partial \cQ}{\partial\bY}}_\rF
    \norm{\ASA}_\rF\norm{\bW_V}_\rF\norm{\bgamma}_\rF;
\end{align*}
\begin{align*}
    \norm{\frac{\partial \cQ}{\partial\bgamma}}_\rF
    \leq&\norm{\frac{\partial \cQ}{\partial\bY}}_\rF\sqrt{n}\norm{\bX_{\rm std}}_\rF\left(\frac{2}{\sqrt{D}}\norm{\Big(\bI_n\otimes\bW_O^\top\bW_V^\top\bX_{\LN}^\top\Big)\frac{\partial\ASA}{\partial\MSA}\left(\bI_n\otimes\bX_{\LN}\bW_K\bW_Q^\top\right)}_\rF+\norm{\ASA\otimes\bW_O^\top\bW_V^\top}_\rF\right)
    \\\leq& n\sqrt{D}\norm{\frac{\partial \cQ}{\partial\bY}}_\rF\left(\frac{2}{\sqrt{D}}
    \norm{\frac{\partial\ASA}{\partial\MSA}}_\rF \norm{\bW_K}_\rF\norm{\bW_Q}_\rF\norm{\bW_V}_\rF\norm{\bW_O}_\rF\norm{\bX_{\LN}}_\rF^2+\norm{\ASA}_\rF\norm{\bW_V}_\rF\norm{\bW_O}_\rF
    \right)
    \\\leq&
    \norm{\frac{\partial \cQ}{\partial\bY}}_\rF\left(\frac{2(n\sqrt{D})^3}{\sqrt{D}}\norm{\frac{\partial\ASA}{\partial\MSA}}_\rF\norm{\bW_K}_\rF\norm{\bW_Q}_\rF\norm{\bW_V}_\rF\norm{\bW_O}_\rF\norm{\bgamma}_\rF^2 + n\sqrt{D}\norm{\ASA}_\rF\norm{\bW_V}_\rF\norm{\bW_O}_\rF\right).
\end{align*}

Therefore, if we define:
\begin{align*}
    \Phi:=&\frac{(n\sqrt{D})^3}{\sqrt{D}}\norm{\frac{\partial \cQ}{\partial\bY}}_\rF\norm{\frac{\partial\ASA}{\partial\MSA}}_\rF\norm{\bW_K}_\rF\norm{\bW_Q}_\rF\norm{\bW_V}_\rF\norm{\bW_O}_\rF\norm{\bgamma}_\rF^3,\\
    \Psi:=&n\sqrt{D}\norm{\frac{\partial \cQ}{\partial\bY}}_\rF\norm{\ASA}_\rF\norm{\bW_V}_\rF\norm{\bW_O}_\rF\norm{\bgamma}_\rF,
\end{align*}
then it holds that:
\begin{gather*}
    \norm{\frac{\partial \cQ}{\partial\bW_{K}}}_\rF\leq\frac{\Phi}{\norm{\bW_K}_\rF};\quad\quad
    \norm{\frac{\partial \cQ}{\partial\bW_{Q}}}_\rF\leq\frac{\Phi}{\norm{\bW_Q}_\rF};
    \\
    \norm{\frac{\partial \cQ}{\partial\bW_{V}}}_\rF\leq\frac{\Psi}{\norm{\bW_V}_\rF};\quad\quad
    \norm{\frac{\partial \cQ}{\partial\bW_{O}}}_\rF\leq\frac{\Psi}{\norm{\bW_O}_\rF};
    \\
    \norm{\frac{\partial \cQ}{\partial\bgamma}}_\rF\leq\frac{2\Phi + \Psi}{\norm{\bgamma}_\rF}.
\end{gather*}

Therefore,
\begin{gather*}
    \cS(\bW_{\bullet})=\frac{1}{\#(\bW_{\bullet})}\norm{\frac{\partial \cQ}{\partial\bW_{\bullet}}}_\rF^2=\cO\left(\frac{\Phi^2}{D^2\norm{\bW_\bullet}_\rF^2}\right),\quad\bullet\in\{K,Q\};
    \\
    \cS(\bW_{\bullet})=\frac{1}{\#(\bW_{\bullet})}\norm{\frac{\partial \cQ}{\partial\bW_{\bullet}}}_\rF^2=\cO\left(\frac{\Psi^2}{D^2\norm{\bW_\bullet}_\rF^2}\right),\quad\bullet\in\{V,O\};
    \\
    \cS(\bgamma)=\frac{1}{\#(\bgamma)}\norm{\frac{\partial \cQ}{\partial\bgamma}}_\rF^2=\cO\left(\frac{\Phi^2+\Psi^2}{D\norm{\bgamma}_\rF^2}\right).
\end{gather*}

\subsection{Proof of Theorem~\ref{thm: Embed vs Norm}}
\label{appendix: proof: thm: Embed vs Norm}

We focus on the transformation from $\bX$ to $\bY:=\LN(\bX\bW_E;\bgamma^{(1/2)})$.
For simplicity, we define:
\begin{align*}
    \bZ:=\bX\bW_E,\quad
    \bZ_{\rm std}:=\frac{\bZ-\bbE_r[\bZ]}{\sqrt{\bbZ_r[\bZ]}},\quad
    \bY=\LN(\bZ;\bgamma)=\bZ_{\rm std}\odot(\bone_{n\times 1}\otimes\bgamma).
\end{align*}

It is straightforward that:
\begin{align*}
    \frac{\partial\bY}{\partial\bgamma}=\diag\big(\vec(\bZ_{\rm std})\big)\big(\bone_{n\times 1}\otimes \bI_{D}\big).
\end{align*}
Recalling the proof in Appendix~\ref{appendix: proof: thm: FFN vs Norm}, we have:
\begin{align*}
    \norm{\frac{\partial\bY}{\partial\bgamma}}_\rF
    \leq n\sqrt{D}.
\end{align*}

Then we calculate $\frac{\partial\bY}{\partial\bW_E}$. For simplicity, we denote 
\begin{gather*}
    \tilde{\bZ}:=\bZ-\bbE_{r}[\bZ],\quad \bZ=\begin{pmatrix}
    \tilde{\bz}_1 \\ ... \\ \tilde{\bz}_d
\end{pmatrix}\in\bbR^{d\times D},\\
\tilde{\bW}_E:=\bW_E-\bbE_{r}[\bW_E],\quad
\bW_E=\begin{pmatrix}
    \bw_{E_1} \\ ... \\ \bw_{E_d}
\end{pmatrix}\in\bbR^{d\times D},\quad
\tilde{\bW}_E=\begin{pmatrix}
    \tilde{\bw}_{E_1} \\ ... \\ \tilde{\bw}_{E_d}
\end{pmatrix}\in\bbR^{d\times D}
\end{gather*} 

By the proof in~\cite{xiong2020layer}, for a vector $\bx\in\bbR^{1\times D}$, denoted by $\tilde{\bx}:=\bx-\bbE[\bx]$, then $\frac{\partial \bx_{\rm std}}{\partial \bx}=\frac{\sqrt{D}}{\norm{\tilde{\bx}}_2}\left(\bI-\frac{\tilde{\bx}^\top\tilde{\bx}}{\norm{\tilde{\bx}}_2^2}\right)\left(\bI-\frac{1}{d}\bone_{1\times D}^\top\bone_{1\times D}\right)$. Thus, we have:
\begin{align*}
    &\frac{\partial\bY}{\partial\bW_E}=\frac{\partial\bY
    }{\partial \bZ_{\rm std}}\frac{\partial \bZ_{\rm std}}{\partial\bZ}\frac{\partial \bZ}{\partial\bW_E}
    \\=&\left(\bI_n\otimes\diag\left(\vec(\bgamma)\right)\right)\diag\left(\left\{\frac{\sqrt{D}}{\norm{\tilde{\bz}_{i}}_2}\left(\bI-\frac{\tilde{\bz}_{i}^\top\tilde{\bz}_{i}}{\norm{\tilde{\bz}_{i}}_2^2}\right)\left(\bI-\frac{1}{D}\bone_{1\times D}^\top\bone_{1\times D}\right)\right\}_{i\in[n]}\right)
    \left(\bX\otimes\bI_D\right).
\end{align*}

Recalling the relationship $z_{i,j}=\sum_{k=1}^d x_{i,k} w_{k,j}$, we have $\bbE[\bz_i]=\sum_{k=1}^d x_{i,k}\bbE[\bw_k]$, which implies
\begin{align*}
    \tilde{\bz}_i=\sum_{k=1}^d x_{i,k} \tilde{\bw}_k.
\end{align*}

Combining this property with the that are one-hot fact of the inputs $\bX$, we have:
\begin{align*}
    \min_{i\in[n]}\norm{\tilde{\bz}_{i}}_2\geq \min_{k\in[d]}\norm{\tilde{\bw}_{k}}_2.
\end{align*}

Additionally, the one-hot encoding ensures:
\begin{align*}
    \norm{\bX}_\rF=\left(\sum_{i=1}^n x_{i,j}^2\right)^{1/2}=\sqrt{n}.
\end{align*}

Now we have the following bound:
\begin{align*}
    &\norm{\frac{\partial\bY}{\partial\bW_E}}_\rF
    \\\leq&\norm{\bI_n\otimes\diag\left(\vec(\bgamma)\right)}_\rF\norm{\diag\left(\left\{\frac{\sqrt{D}}{\norm{\tilde{\bz}_{i}}_2}\left(\bI-\frac{\tilde{\bz}_{i}^\top\tilde{\bz}_{i}}{\norm{\tilde{\bz}_{i}}_2^2}\right)\left(\bI-\frac{1}{D}\bone_{1\times D}^\top\bone_{1\times D}\right)\right\}_{i\in[n]}\right)}_2\norm{\bX\otimes\bI_D}_2
    \\\leq& \sqrt{n}\norm{\bgamma}_\rF\frac{\sqrt{D}}{\min_{i\in[n]}\norm{\tilde{\bz}_i}_2}\norm{\bX}_2\leq n\sqrt{D}\frac{\norm{\bgamma}_\rF}{\min_{i\in[n]}\norm{\tilde{\bz}_i}_2}
    \leq n\sqrt{D}\frac{\norm{\bgamma}_\rF}{\min_{i\in[d]}\norm{\tilde{\bw}_i}_2}.
\end{align*}

If we choose $\Psi:=n\sqrt{D}\norm{\bgamma}_\rF $, then we have:
\begin{align*}
    \norm{\frac{\partial\bY}{\partial\bgamma}}_\rF\leq\frac{\Psi}{\norm{\bgamma}_\rF}
    ,\quad
    \norm{\frac{\partial\bY}{\partial\bW_E}}_\rF\leq\frac{\Psi}{\min_{i\in[d]}\norm{\tilde{\bw}_i}_2}.
\end{align*}

Therefore,
\begin{gather*}
    \cS(\bW_E)=\frac{1}{\#(\bW_E)}\norm{\frac{\partial \cQ}{\partial\bW_E}}_\rF^2=\cO\left(\frac{\Psi^2}{Dd\min_{i\in[d]}\norm{\tilde{\bw}_i}_2^2}\right);
    \\
    \cS(\bgamma)=\frac{1}{\#(\bgamma)}\norm{\frac{\partial \cQ}{\partial\bgamma}}_\rF^2=\cO\left(\frac{\Psi^2}{D\norm{\bgamma}_\rF^2}\right).
\end{gather*}


\begin{thebibliography}{61}
\providecommand{\natexlab}[1]{#1}
\providecommand{\url}[1]{\texttt{#1}}
\expandafter\ifx\csname urlstyle\endcsname\relax
  \providecommand{\doi}[1]{doi: #1}\else
  \providecommand{\doi}{doi: \begingroup \urlstyle{rm}\Url}\fi

\bibitem[Achiam et~al.(2023)Achiam, Adler, Agarwal, Ahmad, Akkaya, Aleman, Almeida, Altenschmidt, Altman, Anadkat, et~al.]{achiam2023gpt}
Achiam, J., Adler, S., Agarwal, S., Ahmad, L., Akkaya, I., Aleman, F.~L., Almeida, D., Altenschmidt, J., Altman, S., Anadkat, S., et~al.
\newblock {GPT}-4 technical report.
\newblock \emph{arXiv preprint arXiv:2303.08774}, 2023.

\bibitem[Ainslie et~al.(2023)Ainslie, Lee-Thorp, de~Jong, Zemlyanskiy, Lebron, and Sanghai]{ainslie-etal-2023-gqa}
Ainslie, J., Lee-Thorp, J., de~Jong, M., Zemlyanskiy, Y., Lebron, F., and Sanghai, S.
\newblock {GQA}: Training generalized multi-query transformer models from multi-head checkpoints.
\newblock In Bouamor, H., Pino, J., and Bali, K. (eds.), \emph{Proceedings of the 2023 Conference on Empirical Methods in Natural Language Processing}, pp.\  4895--4901, Singapore, December 2023. Association for Computational Linguistics.
\newblock \doi{10.18653/v1/2023.emnlp-main.298}.

\bibitem[Brown et~al.(2020)Brown, Mann, Ryder, Subbiah, Kaplan, Dhariwal, Neelakantan, Shyam, Sastry, Askell, et~al.]{brown2020language}
Brown, T., Mann, B., Ryder, N., Subbiah, M., Kaplan, J.~D., Dhariwal, P., Neelakantan, A., Shyam, P., Sastry, G., Askell, A., et~al.
\newblock Language models are few-shot learners.
\newblock \emph{Advances in neural information processing systems}, 33:\penalty0 1877--1901, 2020.

\bibitem[Chen et~al.(2024)Chen, Liang, Huang, Real, Wang, Pham, Dong, Luong, Hsieh, Lu, et~al.]{chen2024symbolic}
Chen, X., Liang, C., Huang, D., Real, E., Wang, K., Pham, H., Dong, X., Luong, T., Hsieh, C.-J., Lu, Y., et~al.
\newblock Symbolic discovery of optimization algorithms.
\newblock \emph{Advances in Neural Information Processing Systems}, 36, 2024.

\bibitem[Cohen et~al.(2021)Cohen, Kaur, Li, Kolter, and Talwalkar]{cohen2021gradient}
Cohen, J.~M., Kaur, S., Li, Y., Kolter, J.~Z., and Talwalkar, A.
\newblock Gradient descent on neural networks typically occurs at the edge of stability.
\newblock \emph{International Conference on Learning Representations}, 2021.

\bibitem[Cohen et~al.(2022)Cohen, Ghorbani, Krishnan, Agarwal, Medapati, Badura, Suo, Cardoze, Nado, Dahl, et~al.]{cohen2022adaptive}
Cohen, J.~M., Ghorbani, B., Krishnan, S., Agarwal, N., Medapati, S., Badura, M., Suo, D., Cardoze, D., Nado, Z., Dahl, G.~E., et~al.
\newblock Adaptive gradient methods at the edge of stability.
\newblock \emph{arXiv preprint arXiv:2207.14484}, 2022.

\bibitem[Cohen et~al.(2024)Cohen, Damian, Talwalkar, Kolter, and Lee]{cohen2024understanding}
Cohen, J.~M., Damian, A., Talwalkar, A., Kolter, Z., and Lee, J.~D.
\newblock Understanding optimization in deep learning with central flows.
\newblock \emph{arXiv preprint arXiv:2410.24206}, 2024.

\bibitem[Devlin(2018)]{devlin2018bert}
Devlin, J.
\newblock Bert: Pre-training of deep bidirectional transformers for language understanding.
\newblock \emph{arXiv preprint arXiv:1810.04805}, 2018.

\bibitem[Dosovitskiy et~al.(2020)Dosovitskiy, Beyer, Kolesnikov, Weissenborn, Zhai, Unterthiner, Dehghani, Minderer, Heigold, Gelly, et~al.]{dosovitskiy2020image}
Dosovitskiy, A., Beyer, L., Kolesnikov, A., Weissenborn, D., Zhai, X., Unterthiner, T., Dehghani, M., Minderer, M., Heigold, G., Gelly, S., et~al.
\newblock An image is worth 16x16 words: Transformers for image recognition at scale.
\newblock \emph{arXiv preprint arXiv:2010.11929}, 2020.

\bibitem[Everett et~al.(2024)Everett, Xiao, Wortsman, Alemi, Novak, Liu, Gur, Sohl-Dickstein, Kaelbling, Lee, et~al.]{everett2024scaling}
Everett, K., Xiao, L., Wortsman, M., Alemi, A.~A., Novak, R., Liu, P.~J., Gur, I., Sohl-Dickstein, J., Kaelbling, L.~P., Lee, J., et~al.
\newblock Scaling exponents across parameterizations and optimizers.
\newblock \emph{arXiv preprint arXiv:2407.05872}, 2024.

\bibitem[Gao et~al.(2020)Gao, Biderman, Black, Golding, Hoppe, Foster, Phang, He, Thite, Nabeshima, et~al.]{gao2020pile}
Gao, L., Biderman, S., Black, S., Golding, L., Hoppe, T., Foster, C., Phang, J., He, H., Thite, A., Nabeshima, N., et~al.
\newblock The {Pile}: An 800{GB} dataset of diverse text for language modeling.
\newblock \emph{arXiv preprint arXiv:2101.00027}, 2020.

\bibitem[George et~al.(2018)George, Laurent, Bouthillier, Ballas, and Vincent]{george2018fast}
George, T., Laurent, C., Bouthillier, X., Ballas, N., and Vincent, P.
\newblock Fast approximate natural gradient descent in a {K}ronecker-factored eigenbasis.
\newblock \emph{Advances in Neural Information Processing Systems}, 31, 2018.

\bibitem[Gokaslan \& Cohen(2019)Gokaslan and Cohen]{Gokaslan2019OpenWeb}
Gokaslan, A. and Cohen, V.
\newblock Openwebtext corpus.
\newblock \url{http://Skylion007.github.io/OpenWebTextCorpus}, 2019.

\bibitem[Grosse \& Martens(2016)Grosse and Martens]{grosse2016kronecker}
Grosse, R. and Martens, J.
\newblock A {K}ronecker-factored approximate {F}isher matrix for convolution layers.
\newblock In \emph{International Conference on Machine Learning}, pp.\  573--582. PMLR, 2016.

\bibitem[Gu \& Dao(2023)Gu and Dao]{gu2023mamba}
Gu, A. and Dao, T.
\newblock Mamba: Linear-time sequence modeling with selective state spaces.
\newblock \emph{arXiv preprint arXiv:2312.00752}, 2023.

\bibitem[Hoffmann et~al.(2022)Hoffmann, Borgeaud, Mensch, Buchatskaya, Cai, Rutherford, Casas, Hendricks, Welbl, Clark, et~al.]{hoffmann2022training}
Hoffmann, J., Borgeaud, S., Mensch, A., Buchatskaya, E., Cai, T., Rutherford, E., Casas, D. d.~L., Hendricks, L.~A., Welbl, J., Clark, A., et~al.
\newblock Training compute-optimal large language models.
\newblock \emph{arXiv preprint arXiv:2203.15556}, 2022.

\bibitem[Hu et~al.(2024)Hu, Tu, Han, He, Cui, Long, Zheng, Fang, Huang, Zhao, et~al.]{hu2024minicpm}
Hu, S., Tu, Y., Han, X., He, C., Cui, G., Long, X., Zheng, Z., Fang, Y., Huang, Y., Zhao, W., et~al.
\newblock Minicpm: Unveiling the potential of small language models with scalable training strategies.
\newblock \emph{arXiv preprint arXiv:2404.06395}, 2024.

\bibitem[Jastrzebski et~al.(2020)Jastrzebski, Szymczak, Fort, Arpit, Tabor, Cho, and Geras]{Jastrzebski2020The}
Jastrzebski, S., Szymczak, M., Fort, S., Arpit, D., Tabor, J., Cho, K., and Geras, K.
\newblock The break-even point on optimization trajectories of deep neural networks.
\newblock In \emph{International Conference on Learning Representations}, 2020.

\bibitem[Jumper et~al.(2021)Jumper, Evans, Pritzel, Green, Figurnov, Ronneberger, Tunyasuvunakool, Bates, {\v{Z}}{\'\i}dek, Potapenko, et~al.]{jumper2021highly}
Jumper, J., Evans, R., Pritzel, A., Green, T., Figurnov, M., Ronneberger, O., Tunyasuvunakool, K., Bates, R., {\v{Z}}{\'\i}dek, A., Potapenko, A., et~al.
\newblock Highly accurate protein structure prediction with alphafold.
\newblock \emph{nature}, 596\penalty0 (7873):\penalty0 583--589, 2021.

\bibitem[Kaddour(2023)]{kaddour2023minipile}
Kaddour, J.
\newblock The {MiniPile} challenge for data-efficient language models.
\newblock \emph{arXiv preprint arXiv:2304.08442}, 2023.

\bibitem[Karpathy(2022)]{Karpathy2022}
Karpathy, A.
\newblock \text{NanoGPT}.
\newblock \url{https://github.com/karpathy/nanoGPT}, 2022.

\bibitem[Keller et~al.(2024)]{jordan2024muon}
Keller, J. et~al.
\newblock Muon optimizer.
\newblock \url{https://github.com/KellerJordan/Muon?tab=readme-ov-file}, 2024.

\bibitem[Kingma \& Ba(2014)Kingma and Ba]{kingma2014adam}
Kingma, D.~P. and Ba, J.
\newblock {Adam}: A method for stochastic optimization.
\newblock \emph{arXiv preprint arXiv:1412.6980}, 2014.

\bibitem[Kunstner et~al.(2024)Kunstner, Yadav, Milligan, Schmidt, and Bietti]{kunstner2024heavy}
Kunstner, F., Yadav, R., Milligan, A., Schmidt, M., and Bietti, A.
\newblock Heavy-tailed class imbalance and why adam outperforms gradient descent on language models.
\newblock \emph{arXiv preprint arXiv:2402.19449}, 2024.

\bibitem[Lei~Ba et~al.(2016)Lei~Ba, Kiros, and Hinton]{lei2016layer}
Lei~Ba, J., Kiros, J.~R., and Hinton, G.~E.
\newblock Layer normalization.
\newblock \emph{ArXiv e-prints}, pp.\  arXiv--1607, 2016.

\bibitem[Liu et~al.(2024{\natexlab{a}})Liu, Feng, Xue, Wang, Wu, Lu, Zhao, Deng, Zhang, Ruan, et~al.]{liu2024deepseek}
Liu, A., Feng, B., Xue, B., Wang, B., Wu, B., Lu, C., Zhao, C., Deng, C., Zhang, C., Ruan, C., et~al.
\newblock Deepseek-v3 technical report.
\newblock \emph{arXiv preprint arXiv:2412.19437}, 2024{\natexlab{a}}.

\bibitem[Liu et~al.(2024{\natexlab{b}})Liu, Li, Hall, Liang, and Ma]{liu2023sophia}
Liu, H., Li, Z., Hall, D., Liang, P., and Ma, T.
\newblock Sophia: A scalable stochastic second-order optimizer for language model pre-training.
\newblock \emph{International Conference on Learning Representations}, 2024{\natexlab{b}}.

\bibitem[Liu et~al.(2019)Liu, Ott, Goyal, Du, Joshi, Chen, Levy, Lewis, Zettlemoyer, and Stoyanov]{liu2019roberta}
Liu, Y., Ott, M., Goyal, N., Du, J., Joshi, M., Chen, D., Levy, O., Lewis, M., Zettlemoyer, L., and Stoyanov, V.
\newblock Roberta: A robustly optimized bert pretraining approach.
\newblock \emph{arXiv preprint arXiv:1907.11692}, 2019.

\bibitem[Loshchilov \& Hutter(2017)Loshchilov and Hutter]{loshchilov2017decoupled}
Loshchilov, I. and Hutter, F.
\newblock Decoupled weight decay regularization.
\newblock \emph{arXiv preprint arXiv:1711.05101}, 2017.

\bibitem[Martens \& Grosse(2015)Martens and Grosse]{martens2015optimizing}
Martens, J. and Grosse, R.
\newblock Optimizing neural networks with {K}ronecker-factored approximate curvature.
\newblock In \emph{International conference on machine learning}, pp.\  2408--2417. PMLR, 2015.

\bibitem[Mi et~al.(2022)Mi, Shen, Ren, Zhou, Sun, Ji, and Tao]{mi2022make}
Mi, P., Shen, L., Ren, T., Zhou, Y., Sun, X., Ji, R., and Tao, D.
\newblock Make sharpness-aware minimization stronger: A sparsified perturbation approach.
\newblock \emph{Advances in Neural Information Processing Systems}, 35:\penalty0 30950--30962, 2022.

\bibitem[Ormaniec et~al.(2024)Ormaniec, Dangel, and Singh]{ormaniec2024does}
Ormaniec, W., Dangel, F., and Singh, S.~P.
\newblock What does it mean to be a transformer? insights from a theoretical hessian analysis.
\newblock \emph{arXiv preprint arXiv:2410.10986}, 2024.

\bibitem[Pesme \& Flammarion(2023)Pesme and Flammarion]{pesme2023saddle}
Pesme, S. and Flammarion, N.
\newblock Saddle-to-saddle dynamics in diagonal linear networks.
\newblock \emph{Advances in Neural Information Processing Systems}, 2023.

\bibitem[Popel \& Bojar(2018)Popel and Bojar]{popel2018training}
Popel, M. and Bojar, O.
\newblock Training tips for the transformer model.
\newblock \emph{arXiv preprint arXiv:1804.00247}, 2018.

\bibitem[Radford et~al.(2019)Radford, Wu, Child, Luan, Amodei, and Sutskever]{radford2019language}
Radford, A., Wu, J., Child, R., Luan, D., Amodei, D., and Sutskever, I.
\newblock Language models are unsupervised multitask learners.
\newblock \emph{OpenAI blog}, 1\penalty0 (8):\penalty0 9, 2019.

\bibitem[Raffel et~al.(2020)Raffel, Shazeer, Roberts, Lee, Narang, Matena, Zhou, Li, and Liu]{raffel2020exploring}
Raffel, C., Shazeer, N., Roberts, A., Lee, K., Narang, S., Matena, M., Zhou, Y., Li, W., and Liu, P.~J.
\newblock Exploring the limits of transfer learning with a unified text-to-text transformer.
\newblock \emph{The Journal of Machine Learning Research}, 21\penalty0 (1):\penalty0 5485--5551, 2020.

\bibitem[Shin et~al.(2024)Shin, Kim, and Moon]{shin2024initializing}
Shin, K.~Y., Kim, S., and Moon, S.-M.
\newblock Initializing the layer-wise learning rate, 2024.
\newblock URL \url{https://openreview.net/forum?id=mSSi0zYkEA}.

\bibitem[Shoeybi et~al.(2019)Shoeybi, Patwary, Puri, LeGresley, Casper, and Catanzaro]{shoeybi2019megatron}
Shoeybi, M., Patwary, M., Puri, R., LeGresley, P., Casper, J., and Catanzaro, B.
\newblock Megatron-lm: Training multi-billion parameter language models using model parallelism.
\newblock \emph{arXiv preprint arXiv:1909.08053}, 2019.

\bibitem[Song et~al.(2024)Song, Ahn, and Yun]{song2024does}
Song, M., Ahn, K., and Yun, C.
\newblock Does sgd really happen in tiny subspaces?
\newblock \emph{arXiv preprint arXiv:2405.16002}, 2024.

\bibitem[Su et~al.(2024)Su, Ahmed, Lu, Pan, Bo, and Liu]{su2024roformer}
Su, J., Ahmed, M., Lu, Y., Pan, S., Bo, W., and Liu, Y.
\newblock Roformer: Enhanced transformer with rotary position embedding.
\newblock \emph{Neurocomputing}, 568:\penalty0 127063, 2024.

\bibitem[Team et~al.(2023)Team, Anil, Borgeaud, Alayrac, Yu, Soricut, Schalkwyk, Dai, Hauth, Millican, et~al.]{team2023gemini}
Team, G., Anil, R., Borgeaud, S., Alayrac, J.-B., Yu, J., Soricut, R., Schalkwyk, J., Dai, A.~M., Hauth, A., Millican, K., et~al.
\newblock Gemini: a family of highly capable multimodal models.
\newblock \emph{arXiv preprint arXiv:2312.11805}, 2023.

\bibitem[Touvron et~al.(2023)Touvron, Lavril, Izacard, Martinet, Lachaux, Lacroix, Rozi{\`e}re, Goyal, Hambro, Azhar, et~al.]{touvron2023LLaMA}
Touvron, H., Lavril, T., Izacard, G., Martinet, X., Lachaux, M.-A., Lacroix, T., Rozi{\`e}re, B., Goyal, N., Hambro, E., Azhar, F., et~al.
\newblock Llama: Open and efficient foundation language models.
\newblock \emph{arXiv preprint arXiv:2302.13971}, 2023.

\bibitem[Vaswani et~al.(2017)Vaswani, Shazeer, Parmar, Uszkoreit, Jones, Gomez, Kaiser, and Polosukhin]{vaswani2017attention}
Vaswani, A., Shazeer, N., Parmar, N., Uszkoreit, J., Jones, L., Gomez, A.~N., Kaiser, {\L}., and Polosukhin, I.
\newblock Attention is all you need.
\newblock \emph{Advances in neural information processing systems}, 30, 2017.

\bibitem[Vyas et~al.(2024)Vyas, Morwani, Zhao, Shapira, Brandfonbrener, Janson, and Kakade]{vyas2024soap}
Vyas, N., Morwani, D., Zhao, R., Shapira, I., Brandfonbrener, D., Janson, L., and Kakade, S.
\newblock Soap: Improving and stabilizing shampoo using adam.
\newblock \emph{arXiv preprint arXiv:2409.11321}, 2024.

\bibitem[Wang et~al.(2024{\natexlab{a}})Wang, Wang, He, Wang, Huang, Xiong, Li, Wu, et~al.]{wang2024improving}
Wang, M., Wang, J., He, H., Wang, Z., Huang, G., Xiong, F., Li, Z., Wu, L., et~al.
\newblock Improving generalization and convergence by enhancing implicit regularization.
\newblock \emph{Advances in Neural Information Processing Systems}, 2024{\natexlab{a}}.

\bibitem[Wang et~al.(2024{\natexlab{b}})]{wang2024understanding}
Wang, M. et~al.
\newblock Understanding the expressive power and mechanisms of transformer for sequence modeling.
\newblock \emph{Advances in Neural Information Processing Systems}, 2024{\natexlab{b}}.

\bibitem[Wen et~al.(2024)Wen, Li, Wang, Hall, Liang, and Ma]{wen2024understanding}
Wen, K., Li, Z., Wang, J., Hall, D., Liang, P., and Ma, T.
\newblock Understanding warmup-stable-decay learning rates: A river valley loss landscape perspective.
\newblock \emph{arXiv preprint arXiv:2410.05192}, 2024.

\bibitem[Wolf et~al.(2020)Wolf, Debut, Sanh, Chaumond, Delangue, Moi, Cistac, Rault, Louf, Funtowicz, Davison, Shleifer, von Platen, Ma, Jernite, Plu, Xu, Scao, Gugger, Drame, Lhoest, and Rush]{wolf-etal-2020-transformers}
Wolf, T., Debut, L., Sanh, V., Chaumond, J., Delangue, C., Moi, A., Cistac, P., Rault, T., Louf, R., Funtowicz, M., Davison, J., Shleifer, S., von Platen, P., Ma, C., Jernite, Y., Plu, J., Xu, C., Scao, T.~L., Gugger, S., Drame, M., Lhoest, Q., and Rush, A.~M.
\newblock Transformers: State-of-the-art natural language processing.
\newblock In \emph{Proceedings of the 2020 Conference on Empirical Methods in Natural Language Processing: System Demonstrations}, pp.\  38--45, Online, October 2020. Association for Computational Linguistics.

\bibitem[Wu et~al.(2018)Wu, Ma, and E]{wu2018sgd}
Wu, L., Ma, C., and E, W.
\newblock How {SGD} selects the global minima in over-parameterized learning: A dynamical stability perspective.
\newblock \emph{Advances in Neural Information Processing Systems}, 31:\penalty0 8279--8288, 2018.

\bibitem[Xie et~al.(2022)Xie, Zhou, Li, Lin, and Yan]{xie2022adan}
Xie, X., Zhou, P., Li, H., Lin, Z., and Yan, S.
\newblock Adan: Adaptive nesterov momentum algorithm for faster optimizing deep models.
\newblock \emph{arXiv preprint arXiv:2208.06677}, 2022.

\bibitem[Xiong et~al.(2020)Xiong, Yang, He, Zheng, Zheng, Xing, Zhang, Lan, Wang, and Liu]{xiong2020layer}
Xiong, R., Yang, Y., He, D., Zheng, K., Zheng, S., Xing, C., Zhang, H., Lan, Y., Wang, L., and Liu, T.
\newblock On layer normalization in the transformer architecture.
\newblock In \emph{International Conference on Machine Learning}, pp.\  10524--10533. PMLR, 2020.

\bibitem[Yang et~al.(2022)Yang, Hu, Babuschkin, Sidor, Liu, Farhi, Ryder, Pachocki, Chen, and Gao]{yang2022tensor}
Yang, G., Hu, E.~J., Babuschkin, I., Sidor, S., Liu, X., Farhi, D., Ryder, N., Pachocki, J., Chen, W., and Gao, J.
\newblock Tensor programs v: Tuning large neural networks via zero-shot hyperparameter transfer.
\newblock \emph{arXiv preprint arXiv:2203.03466}, 2022.

\bibitem[Yang(2019)]{yang2019xlnet}
Yang, Z.
\newblock Xlnet: Generalized autoregressive pretraining for language understanding.
\newblock \emph{arXiv preprint arXiv:1906.08237}, 2019.

\bibitem[Yuan et~al.(2024)Yuan, Liu, Wu, Zhou, and Gu]{yuan2024mars}
Yuan, H., Liu, Y., Wu, S., Zhou, X., and Gu, Q.
\newblock Mars: Unleashing the power of variance reduction for training large models.
\newblock \emph{arXiv preprint arXiv:2411.10438}, 2024.

\bibitem[Zhang \& Sennrich(2019)Zhang and Sennrich]{zhang2019root}
Zhang, B. and Sennrich, R.
\newblock Root mean square layer normalization.
\newblock \emph{Advances in Neural Information Processing Systems}, 32, 2019.

\bibitem[Zhang et~al.(2020)Zhang, Karimireddy, Veit, Kim, Reddi, Kumar, and Sra]{zhang2020adaptive}
Zhang, J., Karimireddy, S.~P., Veit, A., Kim, S., Reddi, S., Kumar, S., and Sra, S.
\newblock Why are adaptive methods good for attention models?
\newblock \emph{Advances in Neural Information Processing Systems}, 33:\penalty0 15383--15393, 2020.

\bibitem[Zhang et~al.(2024{\natexlab{a}})Zhang, Zeng, Wang, and Lu]{zhang2024tinyllama}
Zhang, P., Zeng, G., Wang, T., and Lu, W.
\newblock Tinyllama: An open-source small language model, 2024{\natexlab{a}}.

\bibitem[Zhang et~al.(2024{\natexlab{b}})Zhang, Chen, Ding, Li, Sun, and Luo]{zhang2024transformers}
Zhang, Y., Chen, C., Ding, T., Li, Z., Sun, R., and Luo, Z.-Q.
\newblock Why transformers need adam: A hessian perspective.
\newblock \emph{arXiv preprint arXiv:2402.16788}, 2024{\natexlab{b}}.

\bibitem[Zhang et~al.(2024{\natexlab{c}})Zhang, Chen, Li, Ding, Wu, Ye, Luo, and Sun]{zhang2024adam}
Zhang, Y., Chen, C., Li, Z., Ding, T., Wu, C., Ye, Y., Luo, Z.-Q., and Sun, R.
\newblock Adam-mini: Use fewer learning rates to gain more.
\newblock \emph{arXiv preprint arXiv:2406.16793}, 2024{\natexlab{c}}.

\bibitem[Zhao et~al.(2024{\natexlab{a}})Zhao, Zhang, Chen, Wang, Anandkumar, and Tian]{zhao2024galore}
Zhao, J., Zhang, Z., Chen, B., Wang, Z., Anandkumar, A., and Tian, Y.
\newblock Galore: Memory-efficient llm training by gradient low-rank projection.
\newblock \emph{arXiv preprint arXiv:2403.03507}, 2024{\natexlab{a}}.

\bibitem[Zhao et~al.(2024{\natexlab{b}})Zhao, Morwani, Brandfonbrener, Vyas, and Kakade]{zhao2024deconstructing}
Zhao, R., Morwani, D., Brandfonbrener, D., Vyas, N., and Kakade, S.
\newblock Deconstructing what makes a good optimizer for language models.
\newblock \emph{arXiv preprint arXiv:2407.07972}, 2024{\natexlab{b}}.

\end{thebibliography}
\end{document}